\documentclass[11pt]{article}

\usepackage[final]{acl}

\usepackage{times}
\usepackage{latexsym}
\usepackage[T1]{fontenc}
\usepackage[utf8]{inputenc}
\usepackage{microtype}
\usepackage{inconsolata}
\usepackage{graphicx}
\usepackage{booktabs}
\usepackage{multirow}
\usepackage{amsmath}
\usepackage{amssymb}
\usepackage{algorithm}
\usepackage{algorithmic}
\usepackage[table]{xcolor}
\usepackage{placeins}
\usepackage{subcaption}

\usepackage{array}
\usepackage{makecell}

\usepackage[most]{tcolorbox}
\usepackage{listings}
\lstdefinestyle{promptstyle}{
    basicstyle=\ttfamily\scriptsize,
    breaklines=true,
    columns=fullflexible,
    keepspaces=true,
    showstringspaces=false
}
\newtcblisting{promptbox}[1]{
    title={#1},
    listing only,
    listing options={style=promptstyle},
    breakable,
    enhanced,
    boxrule=0.35pt,
    arc=1mm,
    left=1mm,
    right=1mm,
    top=1mm,
    bottom=1mm,
    colback=blue!3,
    colframe=blue!45!black,
    colbacktitle=blue!18,
    coltitle=black,
    fonttitle=\bfseries
}
\newtcblisting{strategycodebox}[1]{
    title={#1},
    listing only,
    listing options={style=promptstyle,language=Python},
    breakable,
    enhanced,
    boxrule=0.35pt,
    arc=1mm,
    left=1mm,
    right=1mm,
    top=1mm,
    bottom=1mm,
    colback=teal!3,
    colframe=teal!45!black,
    colbacktitle=teal!18,
    coltitle=black,
    fonttitle=\bfseries
}

\title{MetaPS: Adaptive Selection of Programmatic Strategies via Market Simulations}

\title{MetaPS: Adaptive Programmatic Strategy Selection for Market Agents}

\author{
  {\bf Jiaxiang Chen}$^1$, 
  {\bf Aotian Luo}$^1$, 
  {\bf Zhouyi Zheng}$^1$, 
  {\bf Weiyi Huang}$^1$, 
  {\bf Chi Zhang}$^2$, 
  {\bf Zenglin Xu}$^1$ \\
  $^1$Fudan University
  $^2$TensorPacific
}

\begin{document}
\maketitle

\begin{abstract}
No single market strategy always wins: momentum, mean reversion, risk control,
and event-driven rules can each succeed or fail as market conditions change.
Rather than asking large language models to directly generate market actions,
we study an executable decision paradigm where an agent selects from a library
of programmatic strategies, each implemented as a code module mapping market
observations to actions.
We propose \textbf{MetaPS}, a simulation-guided framework for adaptive programmatic strategy selection. MetaPS rolls out candidate strategies in simulated or backtested markets, identifies states where particular strategies lead to better future outcomes, and converts these state--strategy pairs into supervised fine-tuning data. During inference, the simulator is no longer queried: MetaPS observes only the current market state and candidate strategy context, selects a suitable strategy program, and the selected program produces the final action.
Experiments on multi-stock trading and a controlled goods-exchange sandbox show that MetaPS consistently improves across model scales from 0.8B to 9B parameters. It outperforms fixed-strategy baselines, direct decision-making agents, and prompted API-based LLM agents; in several settings, compact fine-tuned models even surpass stronger API models. These results demonstrate that market simulations can provide scalable and targeted supervision for learning adaptive, interpretable, and executable strategy selection.
\end{abstract}

\section{Introduction}

No single strategy consistently wins in financial and economic markets. 
Momentum rules can exploit trend continuation but fail during reversals; mean-reversion rules can benefit from overreaction but lose under persistent news shocks; risk-control rules can preserve capital during stress but miss upside in expansionary regimes. 
These behaviors are consistent with heavy-tailed returns, volatility clustering, and regime-dependent dynamics \citep{mandelbrot1963variation,cont2001empirical,bouchaud2009markets}. 
Market decision-making is therefore not only about designing strong strategies, but also about deciding \emph{when each strategy should be used}.

Recent learning-based trading systems often formulate this problem as direct action prediction, mapping observations to buy, sell, hold, allocation, or exchange decisions \citep{ozbayoglu2020deep,schulman2017proximal,chen2021decision}. 
LLMs further allow decisions to condition on heterogeneous textual and numerical context, such as news, market summaries, and portfolio states \citep{liu2023fingpt,finmem,xiao2024tradingagents}. 
However, direct generation entangles two separable decisions: selecting the market logic appropriate to the current state and executing the concrete action implied by that logic. 
This coupling can make decisions harder to audit, constrain, and reproduce.

We study a more executable paradigm: adaptive selection over programmatic strategies. 
The agent is given a library of strategies, each implemented as a code module that maps market observations to actions. 
These modules play a role similar to tools or skills in LLM agents \citep{react,toolformer,gorilla,chameleon}: they expose reusable executable behavior, while the learned model serves as a meta-level selector. 
Instead of asking the LLM to directly invent a trade, MetaPS asks it to choose which program should control the next decision; the selected program then produces the final action through a deterministic runtime with feasibility, sizing, and risk checks.

The central challenge is supervision. 
Historical market data does not directly label the best strategy for each state, and the answer may depend on future outcomes. 
MetaPS addresses this by using market simulations and backtesting rollouts as targeted supervision. 
For each as-of state, it rolls out candidate strategy programs, estimates their future outcomes, mines effective state--strategy pairs, and converts them into supervised fine-tuning data. 
Future outcomes are used only to construct training targets and are never shown to the model at inference time.

We propose \textbf{MetaPS} (\emph{Meta Programmatic Selector}), a simulation-guided framework for adaptive programmatic strategy selection. 
MetaPS first builds a ranked candidate strategy context from the current market state, then trains an LLM router with simulation-derived labels to select a strategy program and coarse execution fields. 
During inference, the router observes only the as-of state and ranked strategy context, selects a suitable programmatic strategy, and delegates action generation to the selected executable module.

We evaluate MetaPS on historical multi-stock trading and a controlled goods-exchange sandbox. 
Across model scales from 0.8B to 9B parameters, MetaPS improves over fixed-strategy baselines, direct decision-making agents, and prompted API-based LLM agents. 
In several settings, compact fine-tuned MetaPS models outperform stronger API models, suggesting that targeted simulation-derived supervision can be more effective than general-purpose prompting for strategy-level market decisions.

Our contributions are threefold:
\begin{itemize}
    \item We formulate financial and economic market decision-making as adaptive selection over executable programmatic strategies, separating strategy choice from action execution.

    \item We introduce \textbf{MetaPS}, a simulation-guided framework that derives supervised fine-tuning data from candidate strategy rollouts in market simulations and backtests.

    \item We evaluate MetaPS across stock-trading and goods-exchange environments, showing consistent gains across 0.8B--9B models over fixed-strategy, direct-agent, and API-based prompting baselines.
\end{itemize}

We provide related work in Appendix~\ref{app:related_work} to keep the main text focused on the method and empirical analysis.

\section{Method}
\label{sec:method}

\begin{figure*}[t]
    \centering
    \includegraphics[width=1\linewidth]{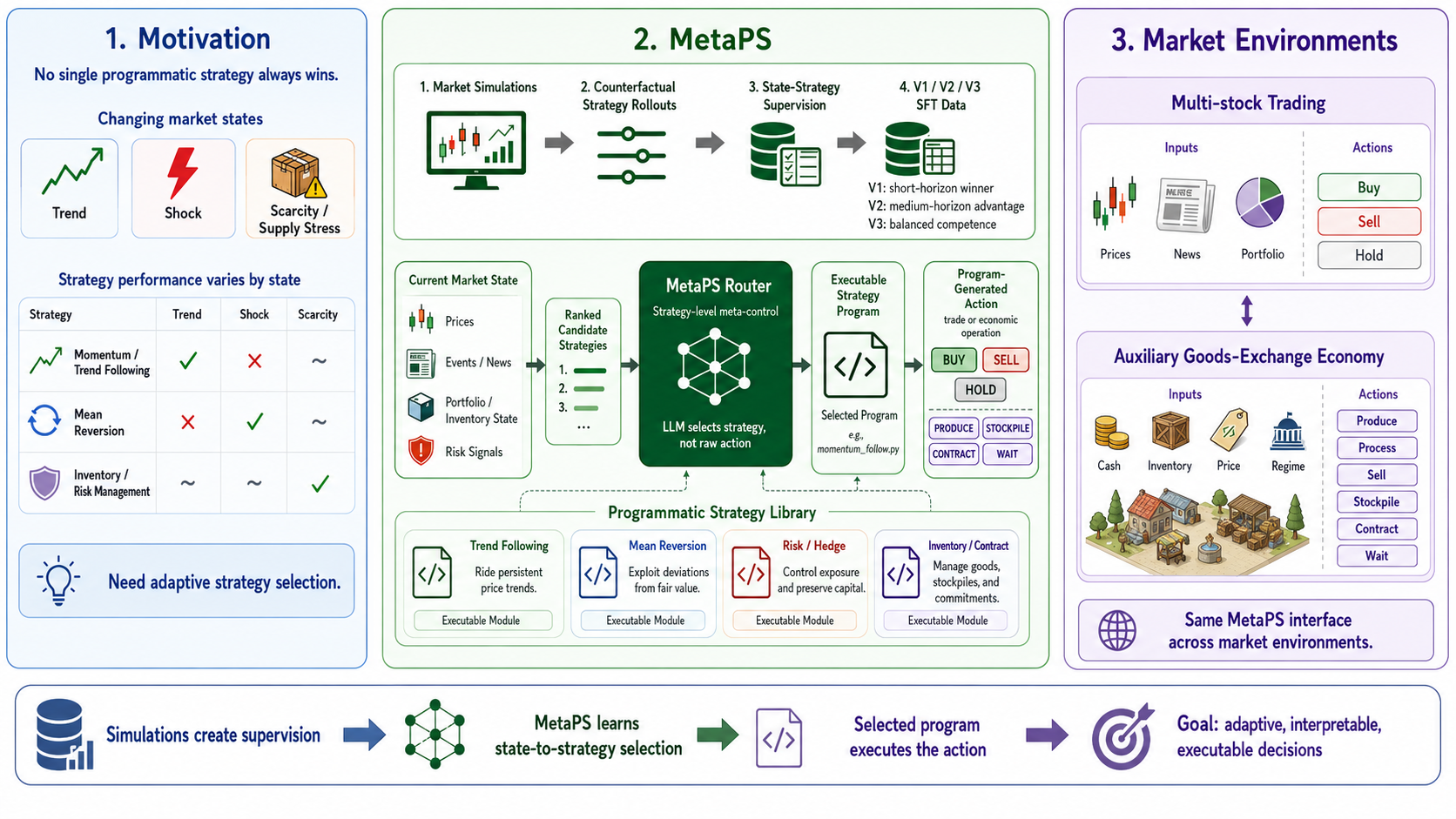}
    \caption{
    Overview of MetaPS. Market simulations supervise a meta-level router by
    rolling out executable strategy programs from the same as-of state. The
    resulting state--strategy winners are converted into supervised fine-tuning
    data. At inference time, MetaPS observes the current market state and a
    ranked candidate set, selects one programmatic strategy, and delegates final
    action generation to the selected executable program.
    }
    \label{fig:metaps_framework}
\end{figure*}
\subsection{Problem Formulation}
\label{subsec:problem_formulation}

We study market decision making with a library of executable programmatic strategies. 
At step $t$, the environment provides an as-of observation $o_t\in\mathcal{O}$, recent history $h_t$, and action space $\mathcal{A}$. 
Rather than mapping observations directly to actions, MetaPS assumes a strategy library
\[
    \mathcal{S}=\{s_1,\ldots,s_N\}, 
    \qquad s_i:\mathcal{O}\rightarrow\mathcal{A},
\]
where each $s_i$ is an executable module that maps the current market state to a feasible action.

MetaPS learns a router over a ranked candidate subset $\mathcal{C}_t\subseteq\mathcal{S}$. 
Given the current observation, recent history, and candidate programs, the router predicts
\[
    z_t=\arg\max_{z\in\mathcal{C}_t}p_\theta(z\mid o_t,h_t,\mathcal{C}_t).
\]
The selected program then produces the final action, $a_t=s_{z_t}(o_t)$. 
Thus, the language model decides \emph{which program to execute}, while the executable strategy decides \emph{which market action to take}. 
This factorization makes the decision process more interpretable, executable, and auditable than direct action generation.

\subsection{Programmatic Strategy Library}
\label{subsec:strategy_library}

The strategy library defines the executable decision primitives available to
MetaPS. Each program follows a shared interface: it consumes an as-of market
state and returns structured action fields, such as direction, target asset or
resource, confidence, risk note, and coarse size. A deterministic execution
layer applies feasibility, sizing, cost, and domain-specific constraints.

The concrete library is domain-specific. In the multi-stock trading environment,
the library covers trend-following, reversal, news-driven, hedging, risk-control,
and breakout-style strategies. In the controlled goods-exchange sandbox, the
library includes producer, processor, contractor, merchant, hoarder, and
balanced programs.
Although the action spaces differ, the MetaPS interface is unchanged: the model
selects a program identifier, and the selected executable module emits the final
action.

\subsection{Simulation-Guided Supervision}
\label{subsec:simulation_learning}

The key challenge is that state--strategy labels are not directly available.
MetaPS constructs them through counterfactual strategy rollouts in a market
simulator or historical backtester $\mathcal{M}$. For an as-of state $o_t$,
$\mathcal{M}$ rolls out each candidate program $s_i\in\mathcal{C}_t$ from the
same state snapshot. With discount factor $\gamma$ and horizon $H$, the rollout
return is
\[
    G_i^{(H)}(o_t)
    =
    \sum_{\tau=0}^{H-1}
    \gamma^\tau r_{t+\tau}^{(i)} ,
\]
where $r_{t+\tau}^{(i)}$ is the reward obtained when actions are generated by
$s_i$ during that counterfactual rollout. A training view $v$ defines a
view-specific score $q_i^v(o_t)$ from these rollout outcomes, and the
corresponding target strategy is
\[
    z_{t,v}^\star
    =
    \arg\max_{s_i\in\mathcal{C}_t}
    q_i^v(o_t).
\]
This produces state--strategy supervision rather than low-level action labels.
The simulations are therefore a training-time source of labels, not an
additional test-time input. Future rollout rewards are used only to construct
labels and are not exposed in the model-facing input.

\paragraph{Candidate strategy context.}
MetaPS does not choose from an unstructured full library. We first construct a
ranked candidate set using a lightweight relevance score $\rho(s_i,o_t)$ and a
candidate budget $B$:
\[
    \mathcal{C}_t
    =
    \operatorname{Top}_{B}
    \{\rho(s_i,o_t)\}_{s_i\in\mathcal{S}} .
\]
The score uses state phrases, strategy triggers, and missing-evidence penalties.
The ranked packet is included in the prompt, keeping the model's choice space
compact while preserving interpretable program options. In the stock
experiments, $N=B=10$, so the ranked context contains the full ten-strategy
library but preserves the scorer's ordering and scores. Appendix~\ref{app:candidate_ranking}
defines the ranking score, and Appendix~\ref{app:ablations} describes the
corresponding input ablations.

\paragraph{Training views.}
We construct a small set of simulation-derived training views
$\mathcal{V}=\{\mathrm{V1},\mathrm{V2},\mathrm{V3}\}$. V1 selects the
short-horizon winner, V2 favors medium-horizon advantage, and V3 balances return
with risk and behavioral regularization. All views share the same input and
output format, but they induce different labels $z_{t,v}^\star$ through
different scores $q_i^v(o_t)$.
They are not a monotonic curriculum; instead, they expose MetaPS to different
strategy-selection biases. Appendix~\ref{app:target_views} gives the exact
scoring definitions.

\subsection{MetaPS Learning and Inference}
\label{subsec:learning_inference}
\label{subsec:sft}
\label{subsec:inference}

The simulation-derived labels are converted into instruction-following
examples. For state $t$ and training view $v\in\mathcal{V}$, the model input
and target text are
\[
    x_t=(o_t,h_t,\mathcal{C}_t),
    \qquad
    y_{t,v}=(z_{t,v}^\star,d_{t,v},m_{t,v}),
\]
where $z_{t,v}^\star$ is the selected strategy label from
Section~\ref{subsec:simulation_learning}, $d_{t,v}$ contains structured
execution fields, and $m_{t,v}$ is a concise rationale. A teacher model rewrites
raw rollout records into fluent rationales while preserving the selected
strategy and action fields. The rewrite is based only on model-visible state
information, so future rewards and counterfactual outcomes do not leak into the
prompt.

Let $\mathcal{D}=\{(x_t,y_{t,v},w_{t,v})\}$ denote the generated SFT data, where
$w_{t,v}$ is an optional confidence weight derived from rollout quality. MetaPS
is trained with a weighted language-model objective:
\[
    \mathcal{L}_{\mathrm{sft}}(\theta)
    =
    -\sum_{(t,v)} w_{t,v} \log p_\theta(y_{t,v}\mid x_t).
\]
The target text teaches the model to produce a strategy-level decision, not a
raw market action. Appendix~\ref{app:sample_weights} details the reward-derived
weights, and Appendix~\ref{app:instruction_format} gives the full response
format.

At inference time, the simulator and rollout rewards are absent. MetaPS receives
only the observable input $x_t=(o_t,h_t,\mathcal{C}_t)$, predicts a strategy
identifier, and then calls the selected executable program:
\[
\begin{aligned}
    \hat z_t
    &= \arg\max_{z\in\mathcal{C}_t} p_\theta(z\mid x_t),\\
    a_t
    &= s_{\hat z_t}(o_t).
\end{aligned}
\]
The environment applies feasibility constraints, sizing rules, costs, slippage,
and domain-specific execution constraints. Outcomes are logged for evaluation;
if the state is later reused for training, they may also become part of a new
simulation-derived supervision set. Additional execution details are provided
in Appendix~\ref{app:execution_details}.

\begin{algorithm}[t]
\small
\caption{MetaPS: simulation-guided programmatic strategy selection}
\label{alg:metaps}
\begin{algorithmic}[1]
\STATE \textbf{Input:} library $\mathcal{S}$, simulator $\mathcal{M}$,
views $\mathcal{V}$
\STATE Initialize SFT dataset $\mathcal{D}\leftarrow\emptyset$
\FOR{each as-of state $o_t$ from $\mathcal{M}$}
    \STATE Rank programs and form $\mathcal{C}_t$
    \FOR{each $s_i\in\mathcal{C}_t$}
        \STATE Roll out $s_i$ and estimate $G_i^{(H)}(o_t)$
    \ENDFOR
    \FOR{each view $v\in\mathcal{V}$}
        \STATE Select label $z_{t,v}^{\star}$ using score $q_i^v(o_t)$
        \STATE Build instruction pair $(x_t,y_{t,v})$
        \STATE Add $(x_t,y_{t,v},w_{t,v})$ to $\mathcal{D}$
    \ENDFOR
\ENDFOR
\STATE Fine-tune $p_\theta$ on $\mathcal{D}$
\STATE At test time, select $\hat z_t$ and execute $s_{\hat z_t}$
\end{algorithmic}
\end{algorithm}

\section{Experiments}
\label{sec:experiments}

The experiments test whether MetaPS improves market decisions by learning
\emph{when to switch among executable strategies}. The primary benchmark is
historical multi-stock trading, where each daily decision uses an as-of market
state and a ranked set of programmatic strategies. We then evaluate the same
interface in a controlled goods-exchange sandbox, analyze the role of ranked
strategy context, compare training objectives across model scales, and test
temporal robustness with expanding-window validation. Additional diagnostics,
including risk--return plots, return curves, prompt formats, and full metrics,
are reported in Appendix~\ref{app:additional_results}.

\subsection{Experimental Setup}
\label{subsec:experimental_setup}

\paragraph{Environments.}
We use two environments with the same high-level strategy-selection interface.
\begin{itemize}
    \item \textbf{Multi-stock trading} is the primary benchmark. It trains on
    2022--2024 and tests on 2025, starts with \$1,000,000, and models
    transaction costs, slippage, and position constraints. Each daily
    observation is strictly as-of and includes price bars, news/event summaries,
    flow, portfolio state, and candidate strategies.
    \item \textbf{Goods-exchange sandbox} is a controlled auxiliary economy
    with cash, inventory, prices, production capacity, contracts, and regime
    notices. It uses the same strategy-selection interface but a non-financial
    execution space.
\end{itemize}
Full environment definitions, metric details, and execution rules are provided
in Appendix~\ref{app:environment_details}.

\paragraph{Programmatic strategy libraries.}
Training and inference use the same executable libraries. The stock library
contains ten strategies, including news impulse, momentum following,
mean-reversion fading, risk reset, macro rotation, and volatility breakout. In
the main Ranked-Strategy setting, all ten stock strategies are visible and
ranked by a lightweight signal scorer; the context therefore contains the full
library while preserving relevance scores and ordering. The sandbox library
contains six economic policies: producer, processor, contractor, merchant,
hoarder, and balanced. In both environments, MetaPS selects \emph{which
strategy should act}; the selected program and runtime produce the final
executable action. Code-level strategy details are in
Appendix~\ref{app:strategy_code_flow}.

\paragraph{Compared systems.}
We compare four method families: strategy-only baselines, non-LLM learned
selectors, LLM prompting baselines, and MetaPS fine-tuned routers. The non-LLM
selectors include classifier, sequence, reward-ranking, and RL policy variants.
LLM baselines use base Qwen or API models without SFT. Unless otherwise stated,
LLM methods use the \emph{Ranked-Strategy} context, where the as-of state is
shown together with ranked strategy scores. For MetaPS, simulations are used
only before training to derive state--strategy supervision; all test-time
methods are evaluated from the same observable state interface.

\subsection{Main Results on Stock Trading}
\label{subsec:main_stock_results}

\begin{table*}[t]
\centering
\small
\caption{Main results across the two evaluation environments. Stock results are
reported on the held-out 2025 benchmark. The goods-exchange sandbox is a
controlled auxiliary environment. Qwen and MetaPS rows report matched-backbone
comparisons at 4B and 9B where available. Higher return, Sharpe, terminal
equity, and winner share are better; lower maximum drawdown is better.}
\label{tab:main_results}
\resizebox{\textwidth}{!}{%
\begin{tabular}{lrrr rrr}
\toprule
& \multicolumn{3}{c}{\textbf{Multi-stock Trading}} &
\multicolumn{3}{c}{\textbf{Goods-exchange Sandbox}} \\
\cmidrule(lr){2-4}\cmidrule(lr){5-7}
Method
& Return $\uparrow$ & Sharpe $\uparrow$ & Max DD $\downarrow$
& Terminal Equity $\uparrow$ & Return $\uparrow$ & Winner Share $\uparrow$ \\
\midrule
\rowcolor{gray!15}\multicolumn{7}{l}{\textbf{Strategy-only Baselines}} \\
Random Strategy & 30.91 & 0.828 & 9.21 & 13401 & 54.84 & 16.2 \\
Best Fixed Strategy & 30.23 & 0.705 & 23.53 & 18609 & 115.02 & 38.0 \\
\addlinespace[1pt]
\rowcolor{gray!15}\multicolumn{7}{l}{\textbf{Non-LLM Learned Selectors}} \\
Strategy Classifier & 35.90 & 0.778 & 16.78 & 26647 & 207.89 & 53.0 \\
Sequence Selector & 19.54 & 0.924 & 18.73 & 23872 & 175.84 & 50.0 \\
Reward-Ranking Selector & 19.64 & 0.275 & 20.87 & 26002 & 200.45 & 53.0 \\
RL Q-Policy Selector & 23.14 & 0.493 & 19.51 & 15030 & 73.67 & 41.0 \\
\addlinespace[1pt]
\rowcolor{gray!15}\multicolumn{7}{l}{\textbf{LLM Prompting Baselines}} \\
Qwen3.5-4B  & 25.12 & 0.916 & 16.59 & 23879 & 175.92 & 41.0 \\
Qwen3.5-9B  & 32.69 & 0.762 & 27.30 & 18912 & 118.53 & 23.0 \\
GPT-5.4-mini & 14.22 & 0.319 & 11.04 & 23249 & 168.64 & 54.0 \\
GPT-5.4 & 32.03 & \textbf{1.112} & \textbf{3.05} & 23554 & 172.16 & 53.0 \\
\addlinespace[1pt]
\rowcolor{green!10}\multicolumn{7}{l}{\textbf{Ours}} \\
MetaPS-4B & 45.72 & 0.705 & 12.57 & \textbf{31514} & \textbf{264.14} & \textbf{59.0} \\
MetaPS-9B & \textbf{50.29} & 0.668 & 15.29 & 30455 & 251.90 & 56.0 \\
\bottomrule
\end{tabular}
}
\end{table*}

\begin{figure}[t]
    \centering
    \includegraphics[width=\linewidth]{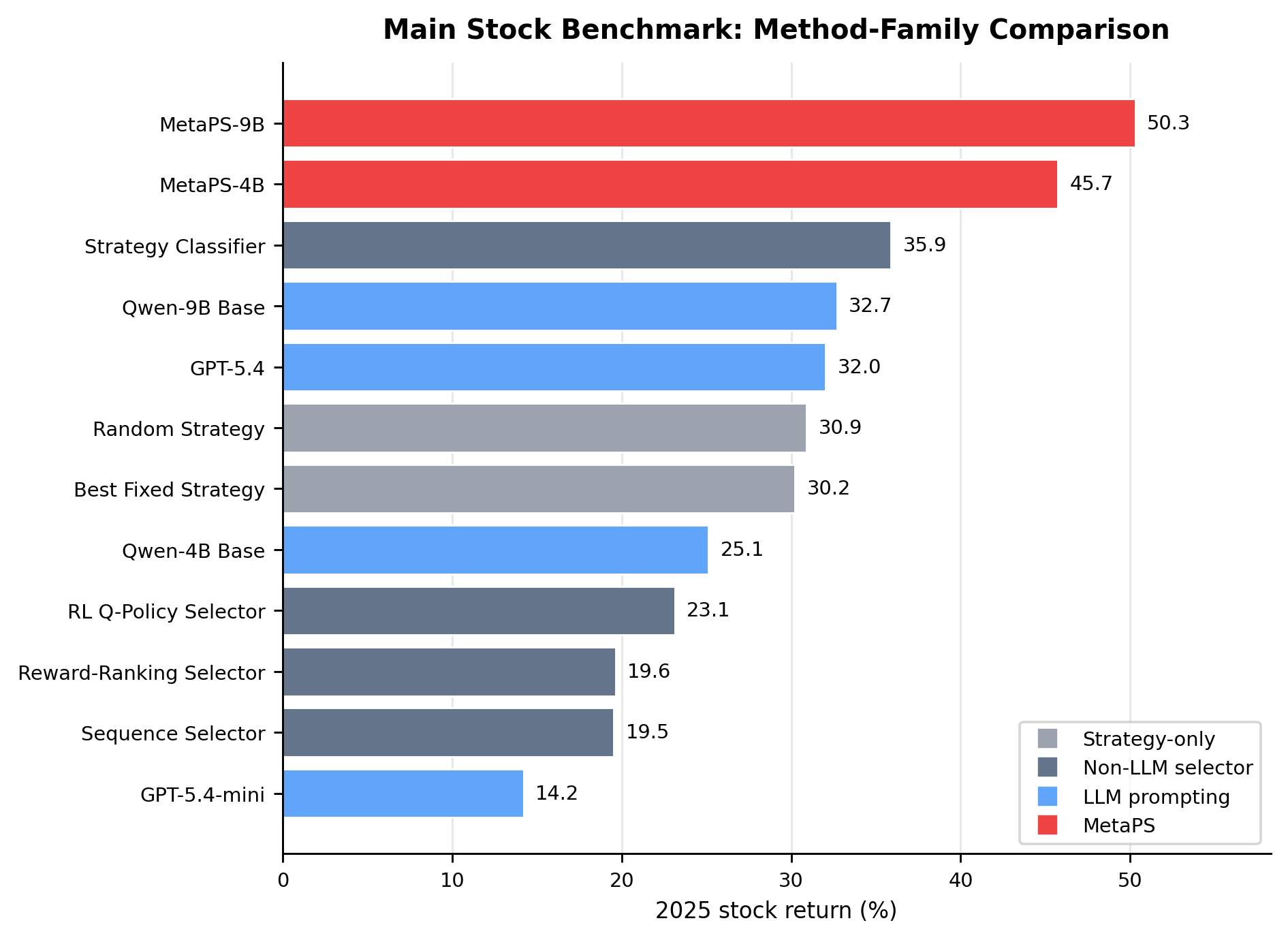}
    \caption{
    Stock-market return comparison across baseline families on the held-out
    2025 benchmark. MetaPS is compared with strategy-only baselines, non-LLM
    learned selectors, base/API LLM prompting, and matched-backbone Qwen
    routers.
    }
    \label{fig:main_baseline_return}
\end{figure}

\begin{figure}[t]
    \centering
    \includegraphics[width=\linewidth]{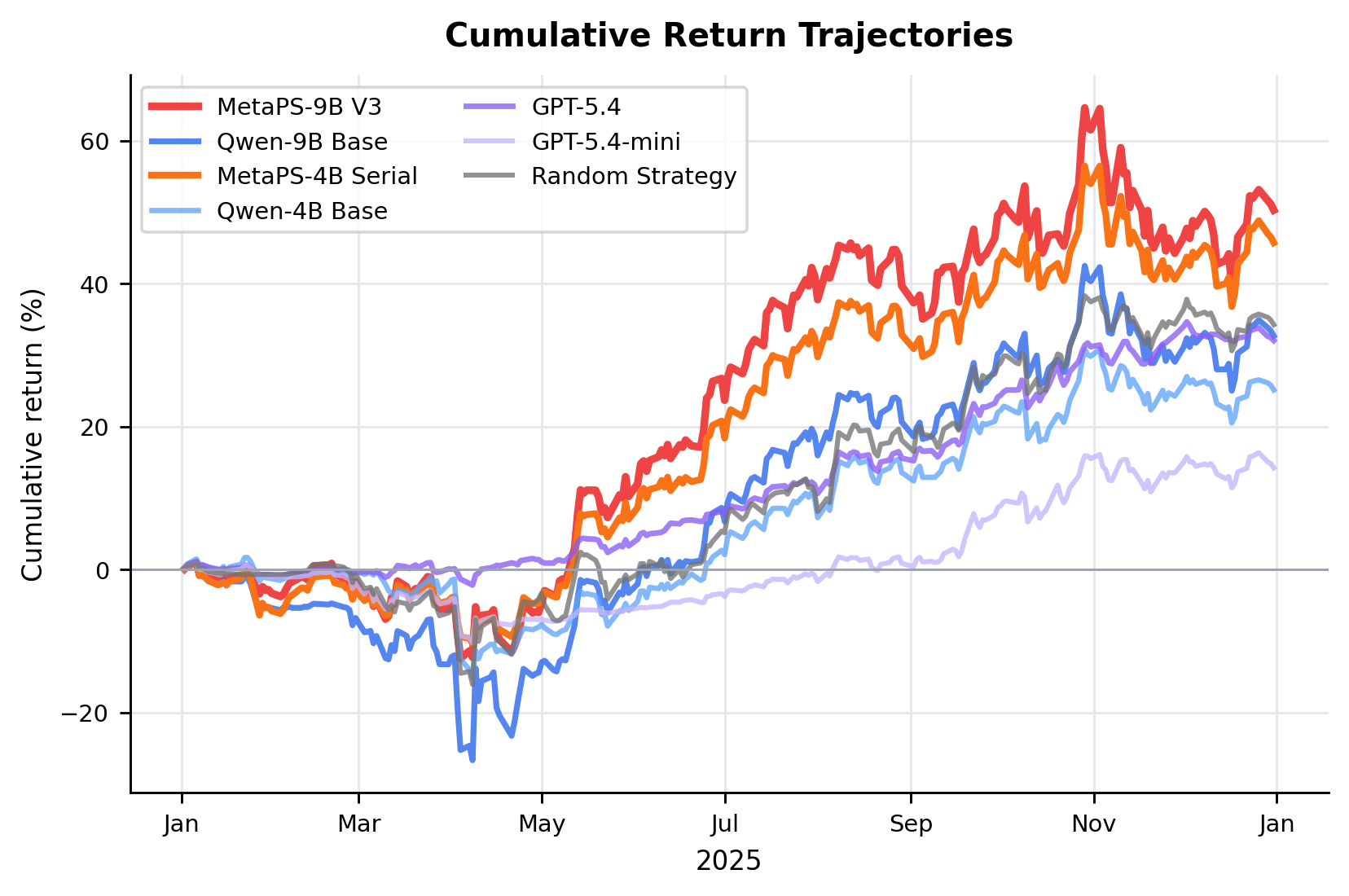}
    \caption{
    Cumulative return trajectories on the held-out 2025 stock benchmark.
    The plot compares MetaPS with Qwen base models, API prompting, and a
    random strategy baseline over the same calendar period.
    }
    \label{fig:return_cumulative_baselines}
\end{figure}

Table~\ref{tab:main_results} reports the main comparison on the held-out 2025 stock benchmark and the controlled sandbox, with Figure~\ref{fig:main_baseline_return} visualizing the stock-return results. 
MetaPS-9B achieves the highest stock return among the main-table methods, reaching 50.29\%, while MetaPS-4B obtains the highest sandbox terminal equity, reaching 31514. 
Matched-backbone comparisons are consistent: MetaPS improves over the Qwen-4B base router in both environments, and the 9B sandbox run increases return from 118.53\% to 251.90\%. 
Since base and fine-tuned routers share the same strategy interface, these gains indicate that supervised meta-policy learning improves strategy switching rather than changing the executable strategy space. 
Figure~\ref{fig:return_cumulative_baselines} further shows that the stock gain is not merely a final-day effect: MetaPS maintains a stronger cumulative-return trajectory through much of the held-out year. 
Additional equity, drawdown, excess-return, monthly, and quarterly diagnostics are provided in Appendix~\ref{app:return_curves}.

The baselines clarify what MetaPS adds. 
Random and fixed strategies achieve non-trivial returns, showing that the library is useful, but their gap to MetaPS indicates that static or unguided strategy use is insufficient. 
Non-LLM selectors, including the retrained classifier, are competitive in some cases but remain below the learned language-based router. 
Prompting-only LLMs offer a complementary comparison: GPT-5.4 achieves the best stock-market Sharpe ratio and lowest drawdown, and both API models transfer to the sandbox without SFT, yet their returns remain below MetaPS in both environments. 
Thus, neither general-purpose prompting nor access to executable strategies alone recovers the high-return, context-dependent routing policy learned by MetaPS.

\subsection{Extension to a Controlled Economic Sandbox}
\label{subsec:economic_sandbox}

We further evaluate MetaPS in a controlled goods-exchange sandbox to test whether the same strategy-routing interface extends beyond historical stock trading. 
The sandbox contains cash, inventory, market price, and six economic regimes: expansion, shortage, processing boom, stable contract, downturn, and volatile markets. 
At each day, the model observes the economic state and selects one of six executable strategies: producer, processor, contractor, merchant, hoarder, or balanced. 
Although the environment differs from ticker-level trading, the learning problem remains unchanged: deciding when to switch among reusable programmatic strategies.

As shown in Table~\ref{tab:main_results}, MetaPS improves over the Qwen base router in this controlled setting. 
For 4B, it raises terminal equity from 23879 to 31514, return from 175.92\% to 264.14\%, and winner share from 41.0\% to 59.0\%; the 9B comparison shows the same trend, more than doubling base return and increasing winner share from 23.0\% to 56.0\%. 
MetaPS also achieves the best terminal equity and winner share among strategy-only, non-LLM, and API prompting baselines. 
We treat the sandbox as a cross-environment extension rather than primary evidence because it is synthetic, but the result suggests that MetaPS is not tied to a specific stock backtester or ticker-level action space. A compact sandbox visualization and full sandbox metrics are reported in Appendix~\ref{app:sandbox_results}.

\subsection{Strategy Behavior Analysis}
\label{subsec:strategy_behavior}

\begin{figure}[t]
    \centering
    \includegraphics[width=0.92\linewidth]{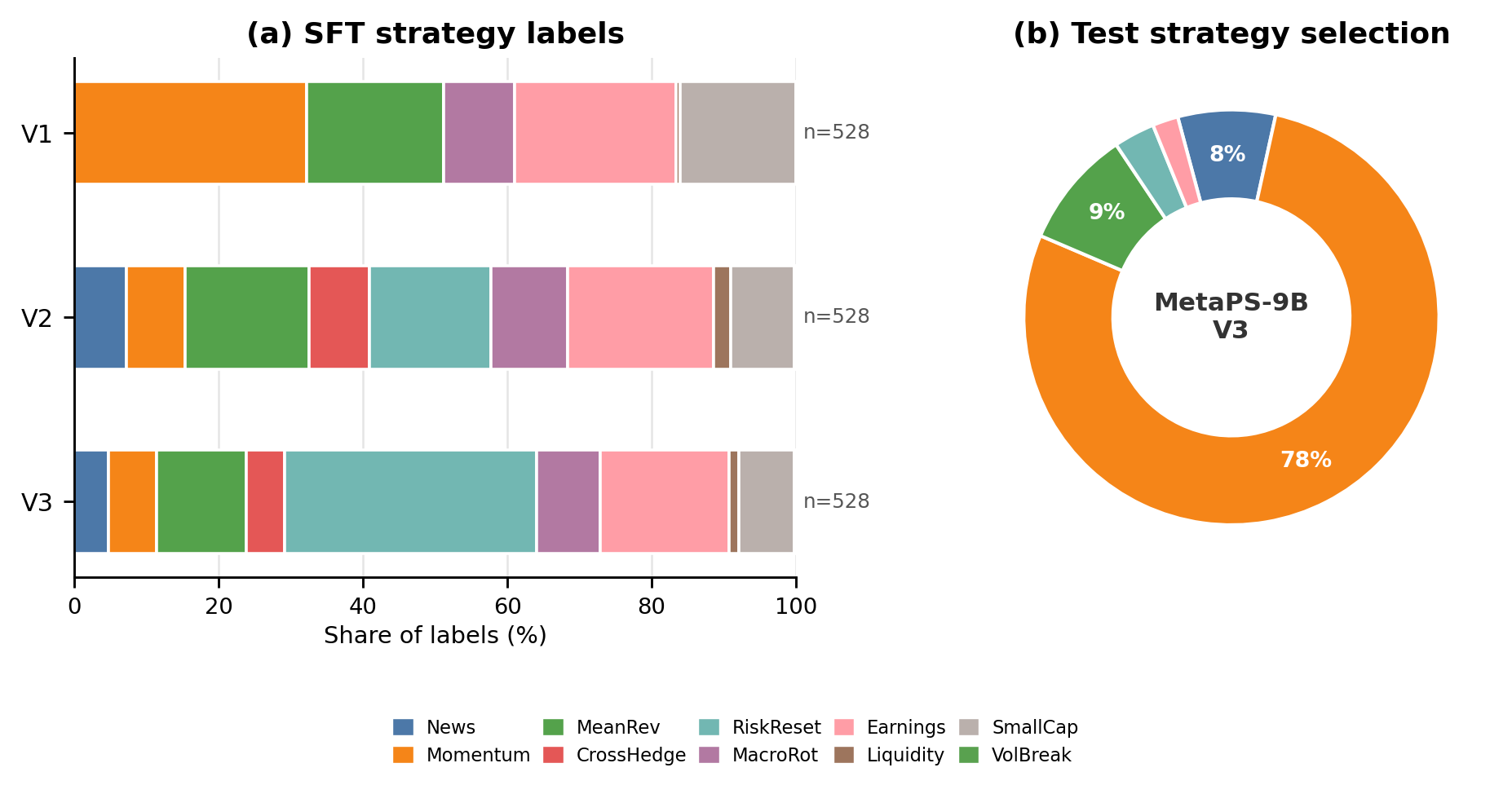}
\caption{
Strategy behavior under simulation-derived supervision. 
Left: strategy-label distributions from the three rollout-based SFT views. 
Right: held-out 2025 strategy-selection distribution of MetaPS-9B V3. 
All ten stock strategies remain available; unused strategies are discussed in Appendix~\ref{app:behavior_diagnostics}.
}
\label{fig:main_strategy_behavior}
\end{figure}

Figure~\ref{fig:main_strategy_behavior} connects the simulation-derived labels
to the router's held-out behavior. The three SFT views correspond to different
ways of converting rollout outcomes into supervision: V1 emphasizes
short-horizon winners, V2 spreads more mass across medium-horizon active
trading strategies, and V3 rebalances toward a broader but still structured
label distribution. At test time, the best overall router does not sample the
library uniformly. In the 2025 regime, MetaPS-9B V3 concentrates on
momentum-following while retaining news impulse, mean-reversion, risk reset,
and earnings drift as secondary programs. This is the intended effect of
simulation guidance: rollouts shape a selective meta-controller over a fixed
library, rather than producing an unconstrained action generator or a uniform
strategy sampler. Cross-scale strategy heatmaps and BUY/HOLD/SELL diagnostics
are reported in Appendix~\ref{app:behavior_diagnostics}.

\begin{figure}
    \centering
    \includegraphics[width=\linewidth]{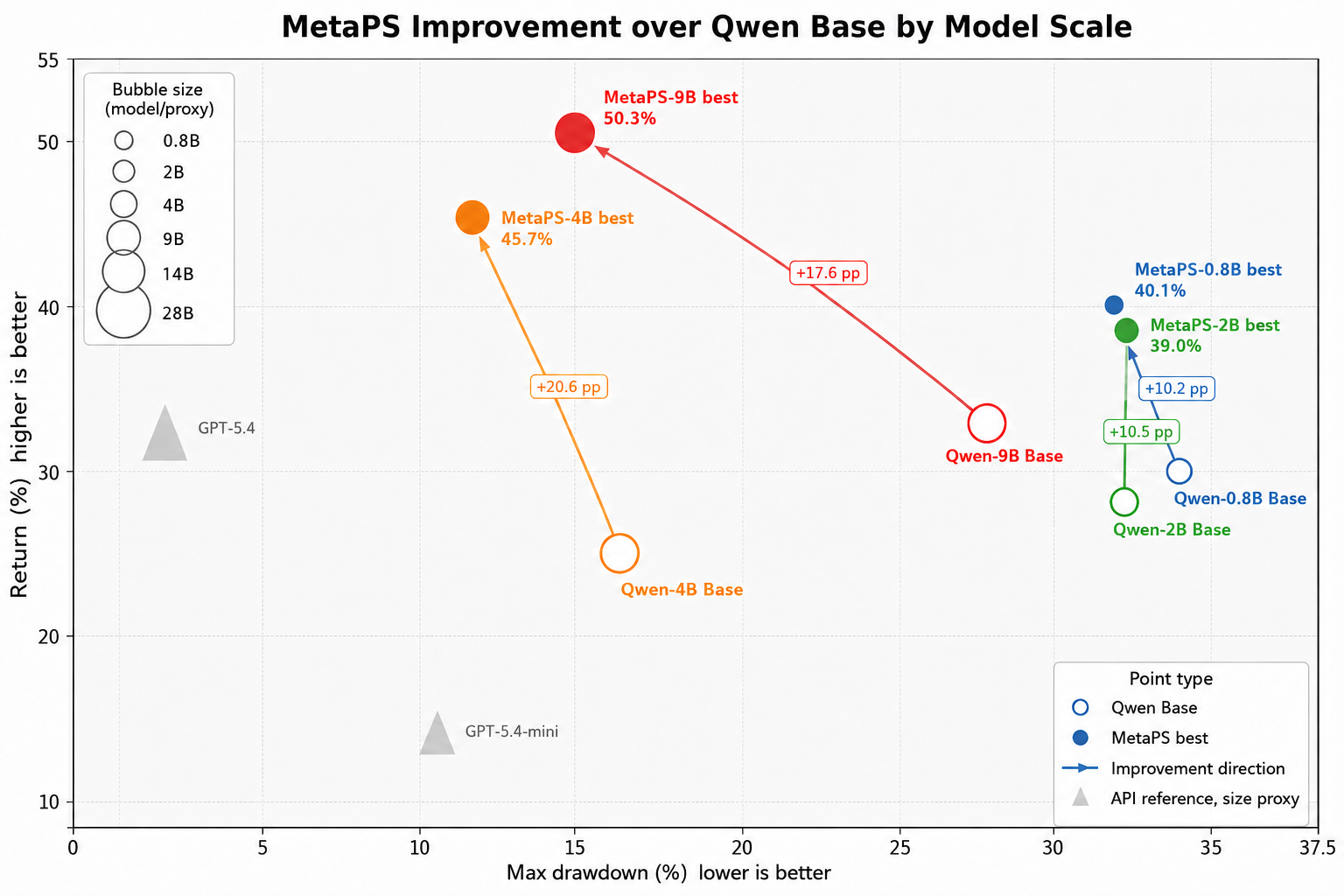}
    \caption{
Risk--return comparison across model scales. 
Hollow circles denote Qwen base models and filled circles denote the best MetaPS variants at matched scales. 
Arrows show that MetaPS consistently shifts the base models toward higher returns, with annotations reporting return gains in percentage points. 
Gray triangles indicate prompting-only API references.
    }
    \label{fig:cmp_across_size}
\end{figure}

\subsection{Effect of Strategy Context}
\label{subsec:strategy_context}

\begin{table}[t]
\centering
\small
\caption{
Ablation of strategy context and action-level prediction on the held-out 2025 stock benchmark.
All Open rows use Qwen3.5-9B; API rows are prompting-only.
}
\label{tab:input_target_ablation}
\setlength{\tabcolsep}{3.5pt}
\renewcommand{\arraystretch}{1.08}
\resizebox{\linewidth}{!}{%
\begin{tabular}{lllcrr}
\toprule
Source & Run & Target & Tuned & Ret. $\uparrow$ & DD $\downarrow$ \\
\midrule

\rowcolor{gray!12}
\multicolumn{6}{l}{\textbf{Ranked-Strategy Context}} \\
API & GPT-5.4 & Strategy & No & 32.03 & \textbf{3.05} \\
API & GPT-5.4-mini & Strategy & No & 14.22 & 11.04 \\
Open & Qwen3.5-9B & Strategy & No & 32.69 & 27.30 \\
Open & MetaPS-9B V3 & Strategy & Yes & \textbf{50.29} & 15.29 \\

\midrule
\rowcolor{gray!12}
\multicolumn{6}{l}{\textbf{All-Strategies Context}} \\
API & GPT-5.4 & Strategy & No & 15.49 & 5.88 \\
API & GPT-5.4-mini & Strategy & No & 11.69 & \textbf{2.62} \\
Open & Qwen3.5-9B  & Strategy & No & 14.51 & 13.91 \\
Open & MetaPS-9B & Strategy & Yes & \textbf{30.12} & 16.60 \\

\midrule
\rowcolor{gray!12}
\multicolumn{6}{l}{\textbf{Direct-Action Context}} \\
API & GPT-5.4 & Action & No & 16.66 & 4.09 \\
API & GPT-5.4-mini & Action & No & 7.71 & 2.38 \\
Open & Qwen3.5-9B & Action & No & 29.30 & 15.14 \\
Open & Qwen3.5-9B SFT & Action & Yes & 3.14 & \textbf{1.65} \\
\bottomrule
\end{tabular}
}
\end{table}

Table~\ref{tab:input_target_ablation} isolates the role of strategy context. 
Ranked-Strategy Context is strongest: MetaPS-9B V3 reaches 50.29\% return, compared with 32.69\% for Qwen3.5-9B under the same context and 30.12\% for MetaPS under All-Strategies Context. 
API-only results show the same pattern: GPT-5.4 performs best with ranked strategies, while unranked or direct-action prompting is weaker. 
This suggests that ranked strategies provide a compact executable decision space, not merely extra text.

The Direct-Action Context shows that the gain is not due to fine-tuning alone. 
With the same 9B backbone and 2022--2024 training window, direct-action SFT reduces return from 29.30\% to 3.14\%, whereas strategy-level SFT reaches 50.29\%. 
This supports the MetaPS design: simulation-derived supervision is more effective for learning strategy-level routing, while the selected program and runtime handle low-level execution. 
Appendix~\ref{app:additional_baseline_figures} and Appendix~\ref{app:direct_action_results} provide the corresponding visualizations and full metrics.

\subsection{Scaling and Objective Analysis}
\label{subsec:scaling_objective}

\begin{table}[t]
\centering
\scriptsize
\caption{Goods-exchange sandbox objective analysis for the 9B router. Values
are reported on the 100-day test split.}
\label{tab:econswitch_9b_objectives}
\setlength{\tabcolsep}{4pt}
\renewcommand{\arraystretch}{1.08}
\resizebox{\linewidth}{!}{%
\begin{tabular}{lrrrrr}
\toprule
Run & Equity $\uparrow$ & Ret. $\uparrow$ & Win. $\uparrow$ & Sharpe $\uparrow$ & DD $\downarrow$ \\
\midrule
Base & 18912 & 118.53 & 23.0 & 14.40 & 1.47 \\
V1 & \textbf{30455} & \textbf{251.90} & 56.0 & 27.01 & \textbf{0.00} \\
V2 & 28705 & 231.68 & 55.0 & 28.24 & \textbf{0.00} \\
V3 & 30302 & 250.13 & \textbf{57.0} & \textbf{34.53} & 1.17 \\
\bottomrule
\end{tabular}
}
\end{table}

\begin{figure*}
    \centering
    \includegraphics[width=0.92\linewidth]{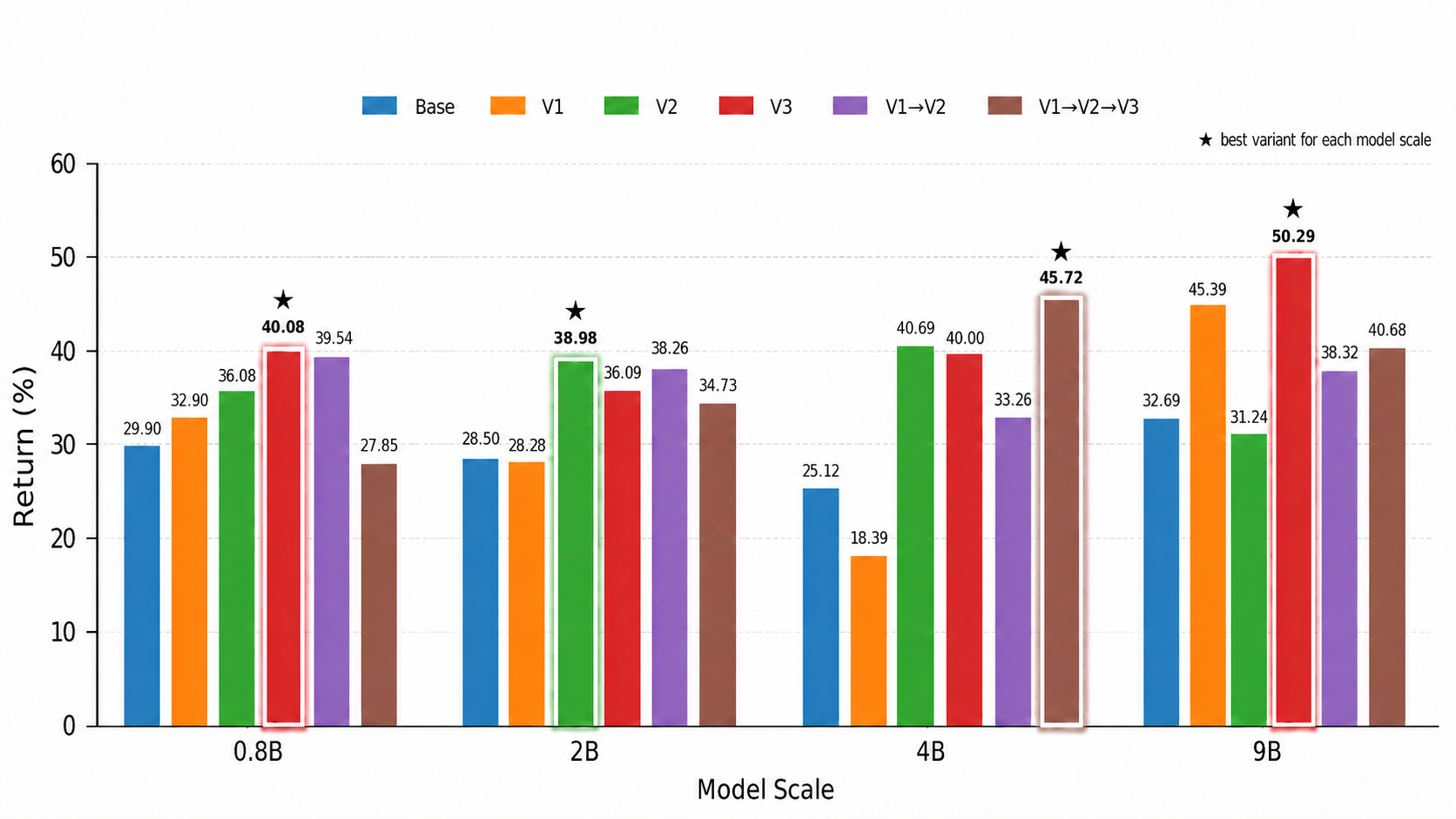}
    \caption{
2025 returns across Qwen model scales under the Ranked-Strategy setting. 
Bars show MetaPS variants, colors denote training objectives, and stars mark the best variant at each scale.
    }
    \label{fig:cmp_across_size_bar}
\end{figure*}

Figure~\ref{fig:cmp_across_size}, and Figure~\ref{fig:cmp_across_size_bar} show that the best training objective is not universal.
V3 is strongest for 0.8B and 9B, V2 for 2B, and the serial V1$\rightarrow$V2$\rightarrow$V3 curriculum for 4B, indicating that V1, V2, and V3 induce different routing behaviors rather than forming a monotonic ladder. 
Across all tested Qwen scales, the best MetaPS variant improves over the corresponding base model: return increases by 10.18 percentage points for 0.8B, 10.48 for 2B, 20.60 for 4B, and 17.60 for 9B. 
The larger 4B and 9B gains are especially favorable because they also reduce maximum drawdown. 
Full metrics for all rows, including final value, Sharpe, drawdown, trades, win rate, and hold rate, are reported in Appendix~\ref{app:full_scale_metrics}; additional heatmap and lift visualizations are in Appendix~\ref{app:additional_baseline_figures}.

Table~\ref{tab:econswitch_9b_objectives} mirrors this objective analysis in the goods-exchange sandbox. 
All three MetaPS objectives substantially improve over the 9B base router, but they emphasize different properties: V1 is best by terminal equity and return, V3 is best by winner share and Sharpe, and V2 provides a middle point with zero observed drawdown. 
This reinforces the interpretation that the objectives encode distinct routing biases rather than a strictly ordered curriculum.

\subsection{Rolling-Time Validation}
\label{subsec:rolling_validation}

The main 2025 benchmark is a fixed train/test split. To test temporal
robustness, we add an expanding-window validation using the 4B router as the
controlled model size. The first split trains on 2022 and tests on
2023--2025. The second trains on 2022--2023 and tests on 2024--2025. The main
setting, 2022--2024 training followed by 2025 testing, is included as the
third split. Because the test horizons differ in length, the table should be
read within each block: it compares objectives under the same train/test
window. This analysis complements the single held-out 2025 benchmark by
checking whether MetaPS remains useful when the training window is shortened
and the test period spans different market regimes.

\begin{table}[t]
\centering
\scriptsize
\caption{Rolling-time validation results of the 4B router. Each block trains on earlier years and tests on subsequent unseen years.}
\label{tab:rolling_4b_full_metrics}
\setlength{\tabcolsep}{4.5pt}
\renewcommand{\arraystretch}{1.08}
\resizebox{\linewidth}{!}{%
\begin{tabular}{lrrrrr}
\toprule
Objective
& Return $\uparrow$
& Trades
& Win Rate $\uparrow$
& Sharpe $\uparrow$
& Max DD $\downarrow$ \\
\midrule
\rowcolor{gray!12}
\multicolumn{6}{l}{\textbf{Train: 2022 \quad Test: 2023--2025}} \\
Base & 94.74 & 305 & 81.48 & \textbf{1.140} & 27.23 \\
V1 & \textbf{197.46} & 63 & \textbf{100.00} & 0.404 & 24.19 \\
V2 & 28.11 & 515 & 58.53 & 0.016 & \textbf{3.29} \\
V3 & 41.20 & 441 & 58.96 & 0.523 & 16.59 \\
V1$\rightarrow$V2 & 55.62 & 275 & 57.72 & -0.073 & 10.87 \\
V1$\rightarrow$V2$\rightarrow$V3 & 128.72 & 279 & 68.38 & -0.007 & 26.79 \\
\addlinespace[2pt]
\rowcolor{gray!12}
\multicolumn{6}{l}{\textbf{Train: 2022--2023 \quad Test: 2024--2025}} \\
Base & 64.46 & 114 & 86.67 & \textbf{0.888} & 29.59 \\
V1 & 100.48 & 67 & 80.77 & -0.020 & \textbf{26.52} \\
V2 & \textbf{117.49} & 100 & 77.42 & 0.046 & 33.66 \\
V3 & 87.39 & 95 & 83.78 & 0.293 & 30.08 \\
V1$\rightarrow$V2 & 113.59 & 86 & \textbf{88.00} & 0.058 & 32.26 \\
V1$\rightarrow$V2$\rightarrow$V3 & 58.06 & 127 & 87.50 & 0.382 & 29.56 \\
\addlinespace[2pt]
\rowcolor{gray!12}
\multicolumn{6}{l}{\textbf{Train: 2022--2024 \quad Test: 2025}} \\
Base & 25.12 & 82 & 32.35 & 0.916 & 16.59 \\
V1 & 18.39 & 76 & 25.00 & \textbf{1.016} & 25.85 \\
V2 & 40.69 & 56 & 28.57 & 0.583 & 28.48 \\
V3 & 40.00 & 88 & \textbf{45.00} & 0.564 & 14.21 \\
V1$\rightarrow$V2 & 33.26 & 54 & 30.77 & 0.581 & 29.47 \\
V1$\rightarrow$V2$\rightarrow$V3 & \textbf{45.72} & 93 & 26.32 & 0.705 & \textbf{12.57} \\
\bottomrule
\end{tabular}
}
\end{table}

Table~\ref{tab:rolling_4b_full_metrics} refines the main 2025 conclusion by testing whether the same objective choices remain useful under shorter training windows and longer unseen horizons. With only 2022
supervision, V1 performs best over 2023--2025, suggesting that short-horizon
winner imitation can identify a small number of high-conviction opportunities.
After adding 2023 to training, V2 becomes strongest over 2024--2025, with the
serial V1$\rightarrow$V2 model close behind. In the shorter 2025-only test, the
full serial curriculum is strongest for 4B and also has the most favorable
drawdown among the 4B trained variants. Thus V3 is not uniformly better or
worse; its value depends on both the market regime and the placement of the
objective in the curriculum. The cumulative 2025 curve in Figure~\ref{fig:return_cumulative_baselines} complements this table by showing the within-year path for the final held-out split.

\section{Conclusion}

We introduced MetaPS, a framework for learning when to switch among interpretable programmatic strategies. 
By shifting learning from low-level market actions to strategy-level meta-control, MetaPS reuses executable strategies while adapting to changing market conditions. 
Results show that strategy routing is effective and supervision choice matters: ranked strategy context outperforms unranked or strategy-free prompting, supervised routing improves over base Qwen models across scales, and rolling-time validation shows that V1, V2, and V3 encode distinct behavioral biases rather than a monotonic curriculum. 
These findings support simulation-guided strategy selection as a practical path toward more executable, interpretable, and adaptive LLM-based market agents.

\section{Limitations}
\label{sec:limitations}

MetaPS is a research framework for strategy selection, not investment advice or a deployable trading system. 
Our stock results are historical backtests with modeled costs, slippage, and constraints, and may not fully capture live-market risks such as liquidity shocks, latency, market impact, or distribution shift. 
The goods-exchange sandbox is synthetic and is used only to test whether the strategy-selection interface transfers beyond stock trading. 
MetaPS also depends on the coverage of its strategy library and the fidelity of its simulator, so missing strategies or biased simulations may affect the learned router. 
Future work should include stronger stress tests, confidence intervals, and live-market robustness checks, with any real-world use requiring human oversight, conservative risk limits, and audit logs.

\bibliography{custom}

\clearpage
\onecolumn
\appendix
\section{Appendix Overview}
\label{app:overview}

The appendix provides implementation and diagnostic details omitted from the
main text: the ranked candidate construction, the three SFT target views,
prompt formats, executable strategy snippets, environment definitions, full
metrics, related work, and additional diagnostic figures.

\section{Related Work}
\label{app:related_work}

\paragraph{Learning-based trading and strategy selection.}
Classical strategy families encode different market assumptions: momentum and
technical rules exploit continuation
\citep{jegadeesh1993returns,brock1992profitability}, reversal effects motivate
contrarian trading \citep{de1985does}, pairs trading captures relative value
\citep{gatev2006pairs}, and volatility targeting adapts exposure to risk
\citep{harvey2018impact}. Learning-based systems instead often map
observations directly to trades, using deep learning, reinforcement learning,
or sequence modeling
\citep{ozbayoglu2020deep,sutton1998reinforcement,schulman2017proximal,chen2021decision,janner2021offline}.
MetaPS is closest to hierarchical control and contextual bandits
\citep{vezhnevets2017feudal,auer2002finite,bubeck2012regret}, but its options
are executable market strategies and its router is trained from simulated
strategy-outcome records.

\paragraph{LLM agents, tools, and skills.}
LLM agents commonly separate reasoning from executable tool use. ReAct links
reasoning traces with actions \citep{react}; Toolformer and related
tool-augmented models study external tool invocation \citep{toolformer,talm};
and Gorilla and BFCL evaluate API/function calling
\citep{gorilla,patil2025bfcl}. Modular and memory-based agents further explore
tool composition and persistent procedural behavior
\citep{chameleon,easytool,generative,memgpt,mem0,memp}. MetaPS applies this
tool/skill view to markets: each tool is an executable strategy, and the LLM
learns when to call it.

\paragraph{Financial LLM agents and market simulation.}
Financial LLMs have been used for financial text processing,
memory-augmented trading, and multi-agent analysis
\citep{liu2023fingpt,finmem,xiao2024tradingagents,lee2024survey,nie2024survey,du2025natural}.
Separately, agent-based market models and simulators study heterogeneous
market dynamics and trading agents
\citep{lux1999scaling,hommes2006heterogeneous,lebaron2006agent,farmer2009economy,byrd2020abides,amrouni2021abides}.
MetaPS combines these directions by using backtesting and simulation not only
for evaluation, but also to generate supervision for a router over executable
strategies.

\section{Additional Method Details}
\label{app:method_details}

This appendix expands the compact method description in
Section~\ref{sec:method}. The main text defines the MetaPS interface and
training objective; here we spell out how the ranked strategy context,
simulation labels, and reward-derived weights are constructed.

\subsection{Notation Map}
\label{app:notation}

The method uses three levels of objects. The \emph{environment level} provides
the observable state $o_t$, history $h_t$, and realized reward $r_t$. The
\emph{strategy level} contains executable programs $s_i\in\mathcal{S}$; each
program maps the current state to an executable action. The \emph{router level}
learns $p_\theta(z\mid o_t,h_t,\mathcal{C}_t)$, where $z$ is a strategy
identifier and $\mathcal{C}_t$ is the ranked candidate set shown to the model.
After the router selects $z_t$, the final market action is produced by the
selected program, $a_t=s_{z_t}(o_t)$.

This separation is important for interpreting the equations below. Rollout
returns and rewards are used to construct supervision for $z_t^\star$; they are
not inserted into the model input at test time. The model-facing input remains
the same for base and fine-tuned models: as-of state, recent feedback, and
ranked strategy candidates.

\subsection{Candidate Ranking}
\label{app:candidate_ranking}

In the ranked-context setting, MetaPS first scores each programmatic strategy
before prompting the language model. The score combines trigger overlap,
textual evidence, a small prior, and missing-evidence penalties:
\[
\begin{split}
    \rho(s_i,o_t) =
    & \alpha \cdot
    |\mathrm{trig}(s_i)\cap\mathrm{phrases}(o_t)| \\
    & + \beta \cdot \mathrm{textmatch}(s_i,o_t)
      + b_i - p_i(o_t).
\end{split}
\]
Here $\mathrm{trig}(s_i)$ denotes phrases associated with strategy $s_i$, while
$\mathrm{phrases}(o_t)$ denotes phrases extracted from the current market state.
The term $\mathrm{textmatch}$ captures additional lexical overlap, $b_i$ is a
small strategy prior, and $p_i(o_t)$ penalizes missing evidence. The top-$K$
strategies under $\rho$ form $\mathcal{C}_t$ in Section~\ref{subsec:simulation_learning}.
This step is not the final policy; it only constructs a compact and
interpretable strategy context for the MetaPS router.

\subsection{Input Ablations}
\label{app:ablations}

The main MetaPS setting uses a ranked top-$K$ strategy context. We compare it
with three simpler alternatives. The \emph{all-strategies} setting exposes the
entire program library without pruning. The \emph{no-strategy} setting removes
program descriptions and asks the model to produce a decision from the market
state alone. The \emph{random-strategy} baseline does not use an LLM router; it
uniformly samples a program from the library and executes it. These ablations
separate the value of programmatic execution from the value of learned adaptive
selection.

\subsection{Simulation-Derived Targets}
\label{app:target_views}

For every as-of state, all candidate strategies in $\mathcal{C}_t$ are evaluated
under the same information set. The three dataset views differ only in how they
turn those simulation outcomes into a target label $z_t^\star$.

The shared pattern is
\[
    z_{t,v}^{\star}
    =
    \arg\max_{s_i\in\mathcal{C}_t}
    q_{i}^{v}(o_t),
\]
where $v\in\{\mathrm{V1},\mathrm{V2},\mathrm{V3}\}$ chooses the scoring rule.
The selected label is then converted into an assistant response
$y_t=(z_t^\star,d_t,m_t)$, where $d_t$ stores execution fields and $m_t$ stores
the rationale.

\paragraph{V1: short-horizon winner.}
V1 directly imitates the current short-horizon simulation winner. If
$G_i^{(1)}(o_t)$ is the one-step or local rollout gain of strategy $s_i$, the
target is
\[
    z_{t,\mathrm{V1}}^\star
    =
    \arg\max_{s_i\in\mathcal{C}_t} G_i^{(1)}(o_t).
\]
This view favors immediate opportunities and high-conviction local actions.
It is closest to a direct imitation of the best local rollout.

\paragraph{V2: medium-horizon advantage.}
V2 evaluates each candidate over multiple horizons
$\mathcal{H}=\{3,5,10,20\}$. For a strategy-specific horizon weight
$\lambda_{i,h}$, the gross long-horizon return is
\[
    R_i^{\mathrm{long}}(o_t)
    =
    \sum_{h\in\mathcal{H}}\lambda_{i,h}G_i^{(h)}(o_t).
\]
The implementation subtracts trading cost, risk, and turnover penalties:
\[
    U_i(o_t)
    =
    R_i^{\mathrm{long}}(o_t)
    - c_i(o_t)-q_i(o_t)-\tau_i(o_t),
\]
where $c_i$ is a transaction-cost proxy, $q_i$ penalizes risky medium/large
positions under risk phrases, and $\tau_i$ penalizes large turnover. The final
score blends this utility with a strategy prior and the candidate's as-of edge:
\[
    B_i(o_t)
    =
    (1-\eta)U_i(o_t)
    + \eta \bar U_i
    + \kappa e_i(o_t),
\]
where $\bar U_i$ is the average utility prior for strategy $s_i$ and
$e_i(o_t)$ is the as-of candidate edge. V2 selects
\[
    z_{t,\mathrm{V2}}^\star
    =
    \arg\max_{s_i\in\mathcal{C}_t} B_i(o_t).
\]
If the best score does not clear the trade margin, the target is converted to a
hold/risk-reset decision. Thus V2 is still supervised learning, but its labels
prefer strategies that remain useful beyond the next step.

\paragraph{V3: balanced competence.}
V3 does not introduce a separate reinforcement-learning objective. Instead, it
constructs a balanced SFT set by choosing, for each state, between the V1 and
V2 labels. Each candidate label receives a quality score
\[
    Q(y_t)
    =
    a(y_t)
    + 3\max(0,B(y_t))
    + 0.08\min(w_t,1.6),
\]
where $a(y_t)$ gives a small active-action bonus, $B(y_t)$ is the V2-style
long-horizon score when available, and $w_t$ is the sample weight. The
selection score then adds distribution-control terms:
\[
    \mathrm{Bal}(y_t)
    =
    Q(y_t)
    + \Delta_{\mathrm{act}}
    + \Delta_{\mathrm{bucket}}
    + \Delta_{\mathrm{strategy}},
\]
where the $\Delta$ terms encourage the dataset to stay near target action,
size-bucket, and strategy frequencies. Labels that exceed their target
frequency are penalized. This is why V3 is best viewed as
return-aware behavioral balancing rather than a pure reward maximization stage.
In short, V1 asks ``what wins now?'', V2 asks ``what remains good over several
steps?'', and V3 asks ``which of the good labels produces a healthier training
distribution?'' These three views all train the same router format from
Section~\ref{subsec:simulation_learning}.

\paragraph{Parallel and serial variants.}
The parallel variants train separate adapters on V1, V2, or V3 data. The serial
variants continue training one adapter on the next dataset, such as
V1$\rightarrow$V2 or V1$\rightarrow$V2$\rightarrow$V3. Thus the serial setting
tests whether the behavioral bias from one view can be refined by later views,
while the parallel setting tests each view as a standalone supervision source.

\subsection{Sample Weights}
\label{app:sample_weights}

Simulation reward affects training through target selection and optional
example weights. The SFT loss is
\[
    \mathcal{L}_{\mathrm{sft}}
    =
    -\sum_t w_t \log p_\theta(y_t\mid x_t).
\]
For long-horizon active decisions, the implementation increases $w_t$ with the
blended score and clips it to a bounded interval. Hold decisions receive a
smaller weight. In the V3 dataset, weights are recomputed from the balanced
quality score and clipped again. The reward values are stored as metadata for
construction and analysis; they are not inserted into the user prompt.
This design keeps inference identical across base and fine-tuned models: at
test time the model only sees the market state and ranked strategy context, not
future reward information.

For tokenized targets $y_t=(y_{t,1},\ldots,y_{t,L_t})$, the log-likelihood term
used above expands to
\[
    \log p_\theta(y_t\mid x_t)
    =
    \sum_{\ell=1}^{L_t}
    \log p_\theta
    (y_{t,\ell}\mid x_t,y_{t,<\ell}).
\]
Thus sample weights scale the whole assistant response for one state--strategy
example, including the selected strategy identifier, execution fields, and
rationale.

\subsection{Instruction Format}
\label{app:instruction_format}

Each SFT example is an instruction-following conversation. The user message
contains the as-of state, the ranked candidate strategies, and brief strategy
descriptions. The assistant message contains fixed Markdown sections:
\emph{Strategy Decision}, \emph{Market Read}, \emph{Why This Strategy},
\emph{Trade Plan}, and \emph{Risk Check}. The selected strategy and action
fields are fixed before teacher rewriting, so the teacher improves rationale
quality without changing the target label.

The system message used for both SFT construction and local-model inference is
short and task-specific. The following boxes show the concrete input and output
schema used by the router. Long market tables and news lists are abbreviated
with ellipses, but the fields and Markdown sections match the data files.

\begin{promptbox}{SFT input prompt template}
SYSTEM:
You are MetaPS, a strategy-routing trading agent. Read the as-of
market context, choose the best candidate strategy, and answer with
the exact Markdown sections requested. Do not use future outcomes or rewards.

USER:
# Strategy Routing Task

Choose which candidate strategy should dominate this as-of market state,
then produce the requested Markdown answer.
Use only the context below. Do not use future rewards, future prices,
or realized outcomes.

## As-Of State
- episode_id=episode-895423
- round_idx=282
- timestamp=16:00
- role_name=strategy.momentum_follow
- env_policy=strategy

## Market Windows
| ticker | price | 1d | 5d | 20d | 60d | points |
| AAPL | 153.71 | -1.04% | +1.88% | +13.63% | +1.60% | 80 |
| NVDA | 22.00 | -3.35% | -1.50% | +31.24% | +42.79% | 80 |
...

## News Context
- round=282 ... headline=No Summons For Elon Musk But Apple's Tim Cook...
- round=282 ... headline=Is the Whole Country Buying Teslas...

## Portfolio State
- cash=...
- gross_exposure=...
- net_exposure=...

## Ranked Candidate Strategies
1. strategy.risk_reset score=... trigger=risk_overheat
2. strategy.momentum_follow score=... trigger=trend_down
...
\end{promptbox}

\begin{promptbox}{SFT assistant target after teacher rewrite}
## Strategy Decision
selected_strategy: strategy.risk_reset
confidence: 0.56

## Market Read
The portfolio exhibits a heavily one-sided book. Market data shows
broad weakness, while news lacks a fresh high-conviction catalyst.

## Why This Strategy
strategy.risk_reset dominates because exposure and volatility risks
matter more than adding another directional trade. Momentum is present,
but the risk_overheat signal makes aggressive following fragile.

## Trade Plan
action: HOLD
ticker: NONE
size_bucket: none

## Risk Check
Do not add risk when constraints dominate; preserve capital before
expressing a new thesis.
\end{promptbox}

\begin{promptbox}{Test-time inference input and parsed output}
INPUT:
Same system and user template as SFT, but with the held-out 2025
as-of state. Example fields: date=2025-01-02, cash=1000000.00,
gross_exposure=0.00, net_exposure=0.0000, volatility_regime=0.1600,
visible price windows, news/event summaries, portfolio state, and
ranked candidate strategies. Future prices, realized P&L, and rollout
rewards are not included.

PARSED OUTPUT:
selected_strategy = strategy.momentum_follow
action            = BUY
ticker            = NVDA
size_bucket       = small
qty               = 289
confidence        = 0.52
\end{promptbox}

Future rewards and realized outcomes remain metadata for dataset construction;
they are not shown in the model prompt. Thus the SFT and inference inputs share
the same observable information, while the SFT target additionally provides the
teacher-rewritten rationale and fixed executable fields.

\begin{table}[h]
\centering
\small
\caption{Data fields used by MetaPS during SFT construction and held-out
inference.}
\label{tab:prompt_data_fields}
\begin{tabular}{p{0.18\linewidth}p{0.36\linewidth}p{0.36\linewidth}}
\toprule
Record & Input content & Output content \\
\midrule
SFT example
& System instruction; as-of state; market windows; news context; portfolio
state; ranked strategy candidates; strategy descriptions.
& Teacher-rewritten Markdown rationale plus parseable fields:
\texttt{selected\_strategy}, \texttt{confidence}, \texttt{action},
\texttt{ticker}, and \texttt{size\_bucket}. Reward and loss weight are stored
only as metadata. \\
\addlinespace[2pt]
Inference trace
& Same observable prompt schema on held-out 2025 states. No future returns,
rollout rewards, or realized P\&L are included.
& Model-generated Markdown response, parsed into the selected strategy and
execution fields. The runtime converts the size bucket to quantity and checks
feasibility before execution. \\
\bottomrule
\end{tabular}
\end{table}

\subsection{Programmatic Strategy Code Flow}
\label{app:strategy_code_flow}

The codebase implements the strategy library as executable modules rather than
as natural-language labels. In the stock environment, each entry in the
\texttt{StockStrategySpec} registry stores a stable strategy key, category,
objective, trigger condition, play description, guardrail, trigger phrases, and
executor path. The roster then creates one participant per strategy, with
\texttt{agent\_id == role\_name == selected\_strategy}. This convention keeps
rollout logs and SFT labels direct: there is no hidden mapping between an
``agent'' and a strategy.

At runtime, \texttt{StockStrategyAgent} dynamically imports the corresponding
executor, e.g., \texttt{game.stock\_game.strategies.momentum\_follow:execute}.
The executor emits a raw decision containing action, ticker, confidence,
optional risk notes, and a coarse size bucket. The wrapper validates the action,
normalizes \texttt{WAIT} to \texttt{HOLD}, clips invalid tickers to
\texttt{HOLD}, and attaches metadata such as
\texttt{selected\_strategy}, \texttt{strategy\_slug}, and
\texttt{risk\_note}. The sizing layer maps
\texttt{none/small/medium/large} to deterministic exposure fractions of
$0\%$, $4\%$, $9\%$, and $16\%$, with quantity caps of $0$, $320$, $800$, and
$1200$ shares. Sell orders are further clipped by available inventory when
short selling is disabled.

The goods-exchange sandbox follows the same abstraction with environment-
specific programs. Its roster maps six roles to executable agents:
\texttt{town\_producer}, \texttt{town\_processor}, \texttt{town\_contractor},
\texttt{town\_merchant}, \texttt{town\_hoarder}, and
\texttt{town\_balanced}. These strategies produce, process, fill contracts,
trade market discounts and premiums, stockpile inventory, or maintain balanced
cash and capacity. The sandbox runtime checks inventory, production capacity,
contract availability, and resource feasibility before applying the final
action.

\subsection{Stock Strategy Library: Meaning and Execution Logic}
\label{app:stock_strategy_snippets}

The ten stock strategies are executable Python modules. Below we show compact
code snippets that preserve each module's decision rule and its intended market
meaning. Quantity emitted by a raw strategy can be present in the strategy
result, but the final evaluated quantity is determined by the shared
size-bucket runtime described above.

\begin{strategycodebox}{strategy.news\_impulse: trade fresh news shocks}
def execute(obs):
    news = obs.news
    if not news:
        return HOLD("no catalyst")
    latest = news[0]
    ticker = argmax_abs(latest.impacts)
    impact = latest.impacts[ticker]
    if abs(impact) < 0.002:
        return HOLD("news impact too weak")
    direction = "BUY" if impact > 0 else "SELL"
    return {
        "action": f"{direction} {ticker}",
        "size_mode": "probe_to_medium",
        "confidence": min(abs(impact) * 30, 1.0),
        "risk_note": "abort if follow-through stalls",
    }
\end{strategycodebox}

\begin{strategycodebox}{strategy.momentum\_follow: ride confirmed short-term momentum}
def execute(obs):
    best_ticker, best_momentum = None, 0.0
    for ticker, history in obs.price_history.items():
        move_3 = ret(history[-4], history[-1])
        if abs(move_3) > abs(best_momentum):
            best_ticker, best_momentum = ticker, move_3
    if best_ticker is None or abs(best_momentum) < 0.002:
        return HOLD("momentum too weak")
    direction = "BUY" if best_momentum > 0 else "SELL"
    return {
        "action": f"{direction} {best_ticker}",
        "size_mode": "scalable",
        "confidence": min(abs(best_momentum) * 20, 1.0),
        "risk_note": "pair with hedge if the book becomes one-sided",
    }
\end{strategycodebox}

\begin{strategycodebox}{strategy.mean\_revert\_fade: fade overextended moves}
def execute(obs):
    candidates = []
    for ticker, history in obs.price_history.items():
        sma5 = mean(history[-5:])
        deviation = (history[-1] - sma5) / sma5
        if abs(deviation) > 0.015:
            direction = "SELL" if deviation > 0 else "BUY"
            candidates.append((ticker, direction, abs(deviation)))
    if not candidates:
        return HOLD("no overextension")
    ticker, direction, stretch = select_fade_candidate(candidates)
    return {
        "action": f"{direction} {ticker}",
        "size_mode": "small_probe",
        "confidence": min(stretch / 0.05, 1.0),
        "risk_note": "avoid fading fresh catalysts without exhaustion",
    }
\end{strategycodebox}

\begin{strategycodebox}{strategy.cross\_asset\_hedge: offset a one-sided book}
def execute(obs):
    equity = obs.cash + obs.gross_exposure
    net_ratio = abs(obs.net_exposure) / max(equity, 1.0)
    gross_ratio = obs.gross_exposure / max(equity, 1.0)
    if not (net_ratio > 0.4 and gross_ratio > 0.2):
        return HOLD("book is balanced")
    hedge_direction = infer_hedge_direction(obs.positions)
    hedge_ticker = find_cross_asset_offset(obs.price_history, obs.positions)
    if hedge_ticker is None:
        return HOLD("no suitable hedge instrument")
    action = "SELL" if hedge_direction == "short" else "BUY"
    return {
        "action": f"{action} {hedge_ticker}",
        "size_mode": "counterbalance",
        "risk_note": "hedge legs stay subordinate to the main thesis",
    }
\end{strategycodebox}

\begin{strategycodebox}{strategy.risk\_reset: preserve capital under constraints}
def execute(obs):
    equity = obs.cash + obs.gross_exposure
    vol_risk = max(0, obs.volatility_regime - 0.4) / 0.6
    liq_risk = max(0, obs.liquidity_regime - 0.4) / 0.6
    exposure_risk = obs.gross_exposure / max(equity, 1.0)
    risk = 0.4 * vol_risk + 0.3 * liq_risk + 0.3 * exposure_risk
    if risk < 0.5:
        return HOLD("risk elevated but no forced reset")
    return {
        "action": "REDUCE",
        "size_mode": "reduce",
        "confidence": 0.6 * (1.0 - 0.3 * risk),
        "risk_note": "no new positions until constraints cool",
    }
\end{strategycodebox}

\begin{strategycodebox}{strategy.macro\_rotation: rotate across macro and asset regimes}
def execute(obs):
    macro_signal = news_macro_impact(obs.news)
    macro_signal += cross_asset_momentum_spread(obs.price_history)
    macro_signal += factor_risk_appetite(obs.factor_state)
    if abs(macro_signal) < 0.01:
        return HOLD("no clear cross-asset rotation")
    direction = "risk_on" if macro_signal > 0 else "risk_off"
    ticker = best_asset_for_regime(obs.price_history, direction)
    action = "BUY" if macro_signal > 0 else "SELL"
    return {
        "action": f"{action} {ticker}",
        "size_mode": "balanced",
        "confidence": min(abs(macro_signal), 1.0),
        "risk_note": "require multi-asset confirmation before scaling",
    }
\end{strategycodebox}

\begin{strategycodebox}{strategy.earnings\_drift: follow post-event repricing}
def execute(obs):
    tickers = tickers_mentioned_by_recent_news(obs.news)
    candidates = []
    for ticker in tickers:
        drift = ret(obs.price_history[ticker][-3], obs.price_history[ticker][-1])
        if 0.003 < abs(drift) < 0.06:
            direction = "BUY" if drift > 0 else "SELL"
            candidates.append((ticker, direction, abs(drift)))
    if not candidates:
        return HOLD("wait for post-event drift confirmation")
    ticker, direction, strength = max(candidates, key=lambda x: x[2])
    return {
        "action": f"{direction} {ticker}",
        "size_mode": "balanced",
        "confidence": min(strength / obs.volatility_regime, 1.0),
        "risk_note": "skip broad macro noise",
    }
\end{strategycodebox}

\begin{strategycodebox}{strategy.liquidity\_rebate: provide liquidity in two-way markets}
def execute(obs):
    two_way = obs.liquidity_regime > 0.4 and obs.volatility_regime < 0.5
    if not two_way or obs.news:
        return HOLD("avoid liquidity provision during catalysts")
    candidates = []
    for ticker, history in obs.price_history.items():
        mid = mean(history[-5:])
        spread = abs(history[-1] - mid) / mid
        if spread < 0.008:
            candidates.append((ticker, history[-1], mid, spread))
    if not candidates:
        return HOLD("no tight-spread instrument")
    ticker, price, mid, spread = max(candidates, key=lambda x: x[3])
    action = "BUY" if price < mid else "SELL"
    return {
        "action": f"{action} {ticker}",
        "size_mode": "liquidity",
        "confidence": 0.47,
        "risk_note": "stop when one-way news gaps appear",
    }
\end{strategycodebox}

\begin{strategycodebox}{strategy.small\_cap\_breakout: trade high-beta style expansion}
def execute(obs):
    if obs.liquidity_regime < 0.3:
        return HOLD("liquidity too thin")
    candidates = []
    for ticker, history in obs.price_history.items():
        recent_high, recent_low = max(history[-6:]), min(history[-6:])
        range_pct = (recent_high - recent_low) / recent_low
        vol_expansion = recent_vol(history) / older_vol(history)
        if history[-1] > recent_high * 0.998 and range_pct > 0.01:
            candidates.append((ticker, range_pct * vol_expansion))
    if not candidates:
        return HOLD("no high-beta breakout")
    ticker, strength = max(candidates, key=lambda x: x[1])
    return {
        "action": f"BUY {ticker}",
        "size_mode": "breakout",
        "confidence": min(strength, 1.0),
        "risk_note": "cut size when breadth fades",
    }
\end{strategycodebox}

\begin{strategycodebox}{strategy.volatility\_breakout: trade range breaks after compression}
def execute(obs):
    candidates = []
    for ticker, history in obs.price_history.items():
        window = history[-20:]
        high, low, current = max(window), min(window), history[-1]
        range_width = (high - low) / low
        if range_width >= 0.005:
            continue
        if current > high * 1.002:
            candidates.append((ticker, "BUY", breakout_strength(current, high, low)))
        if current < low * 0.998:
            candidates.append((ticker, "SELL", breakout_strength(current, high, low)))
    if not candidates:
        return HOLD("no compression-breakout pattern")
    ticker, action, strength = max(candidates, key=lambda x: x[2])
    return {
        "action": f"{action} {ticker}",
        "size_mode": "breakout",
        "confidence": min(strength, 1.0),
        "risk_note": "false-break risk is high when liquidity is thin",
    }
\end{strategycodebox}

\subsection{Execution Details}
\label{app:execution_details}

At inference time, MetaPS only selects a strategy program and coarse execution
fields. The selected program is then executed by the environment-specific
runtime. In stock trading, the runtime checks ticker validity, position limits,
cash constraints, transaction costs, slippage, and size-bucket rules. In the
goods-exchange sandbox, it checks inventory, production capacity, order
availability, and resource feasibility. This separation keeps the learned model
at the strategy-selection level while ensuring that final actions remain
executable under the simulator.

\section{Environment Details}
\label{app:environment_details}

This section describes the two environments used in the experiments. Both are
implemented through the same MetaPS interface: the environment constructs an
as-of observation, the router selects a strategy program, and the selected
program produces an executable action checked by the runtime.

\subsection{Multi-stock Trading Environment}
\label{app:stock_env}

The stock benchmark is the primary environment. It uses daily market data and
news/event context for a multi-asset universe consisting of AAPL, NVDA, SPY,
QQQ, GLD, and USO. Each trading day is treated as one decision step. The main
split trains on 2022--2024 and tests on 2025; the rolling validation in
Section~\ref{subsec:rolling_validation} uses earlier expanding-window splits.
The portfolio starts with \$1,000,000 in cash.

\paragraph{Observation.}
At day $t$, the model receives only information available as of that day. The
observation includes recent price bars, short- and medium-horizon return
windows, news/event summaries, recent order flow, cash and position state, and
a ranked strategy context. The ranked context is built from the same ten stock
strategy programs used at inference time. In the Ranked-Strategy setting, all
ten stock programs are visible, ordered by the lightweight strategy scorer.

\paragraph{Action.}
The router emits a structured decision containing a selected strategy, action
(\textsc{buy}, \textsc{sell}, or \textsc{hold}), ticker, confidence, and size
bucket. The runtime converts the size bucket into a deterministic quantity and
then checks cash, position limits, transaction costs, slippage, and ticker
validity. This separates the learned part of the system from low-level
execution mechanics.

\paragraph{Reward and metrics.}
During dataset construction, candidate strategies are rolled out under the same
as-of state to construct state--strategy supervision. During evaluation, we
report portfolio-level metrics: total return, final value, Sharpe ratio,
maximum drawdown, trade count, win rate, and hold rate. Return is computed from
final portfolio value relative to initial cash. Sharpe is computed from the
sequence of daily portfolio returns, and maximum drawdown is the largest
peak-to-trough portfolio decline during the test period.

\subsection{Controlled Goods-exchange Sandbox}
\label{app:sandbox_env}

The goods-exchange sandbox is an auxiliary environment designed to test whether
the MetaPS interface can be instantiated outside historical financial data. It
is intentionally simpler than the stock benchmark. The state contains cash,
inventory, market price, and regime signals. The available strategy programs
are producer, processor, contractor, merchant, hoarder, and balanced. Each strategy maps
the current economic state to a feasible inventory or market action.

The sandbox contains six regimes: expansion, shortage, processing boom, stable
contract, downturn, and volatile markets. These regimes change the relative
value of producing, holding inventory, accepting contracts, and selling goods.
Terminal equity is cash plus inventory value and any simulator-defined asset
value at the end of the episode. Winner share is the percentage of days on
which the selected policy matches the best strategy under the simulator's
same-day evaluation, and mean regret is the average gap to that best available
strategy. Because these dynamics are synthetic, we use the sandbox as a
controlled interface check rather than as the main empirical claim.

\section{Additional Results and Diagnostics}
\label{app:additional_results}

This appendix provides a more detailed view of the empirical results reported in
Section~\ref{sec:experiments}. The goal is not to introduce a separate benchmark,
but to make the main claims easier to audit. We therefore keep the same held-out
2025 stock evaluation wherever possible and add diagnostics that expose three
aspects that are compressed in the main tables: the effect of the prediction
target, the timing of gains and losses, and the behavioral distribution of the
learned routers. The final subsection reports a controlled sandbox experiment as
auxiliary evidence that the same strategy-selection interface can be instantiated
outside the stock backtester.

\subsection{Direct-Action SFT Baseline}
\label{app:direct_action_results}

Table~\ref{tab:direct_action_full} expands the direct-action comparison from
Table~\ref{tab:input_target_ablation}. This ablation removes the strategy
identifier from the supervision target and asks the model to emit the final
action fields directly. It is therefore a stricter and more brittle learning
problem: the model must imitate low-level trading decisions whose quality depends
on the local market path, the current portfolio state, and the interaction
between entry timing and subsequent price movement. In contrast, MetaPS uses a
strategy identifier as an intermediate decision target, so the model learns which
programmatic behavior to route to before the deterministic backtester converts
that choice into concrete trades.

The resulting gap is substantial. The direct-action base model still produces
active trades and reaches a positive terminal return, but direct-action SFT
collapses most of this return while preserving the same trade count and win rate.
The lower drawdown of the SFT model should therefore not be read as a stronger
risk-adjusted policy; in this run it mostly reflects the fact that the model
fails to compound capital rather than a better return--risk trade-off. This
failure mode supports the central design choice of MetaPS: low-level action
labels are noisy and path-dependent, whereas strategy identifiers provide a more
stable and reusable decision abstraction.

\begin{table}[!t]
\centering
\small
\caption{Full direct-action baseline metrics on the held-out 2025 stock
benchmark. Direct-action SFT preserves the number of executed trades but loses
most of the return obtained by the base model, indicating that the degradation is
not simply caused by inactivity.}
\label{tab:direct_action_full}
\setlength{\tabcolsep}{4pt}
\renewcommand{\arraystretch}{1.08}
\resizebox{\linewidth}{!}{%
\begin{tabular}{lrrrrrrr}
\toprule
Method & Return $\uparrow$ & Sharpe $\uparrow$ & Max DD $\downarrow$
& Trades & Win Rate $\uparrow$ & Hold Rate & Final Value $\uparrow$ \\
\midrule
Direct Action Base & 29.30 & 0.077 & 15.14
& 50 & 50.0 & 80.0 & 1292956.86 \\
Direct Action SFT & 3.14 & -0.139 & \textbf{1.65}
& 50 & 50.0 & 80.0 & 1031418.18 \\
\bottomrule
\end{tabular}
}
\end{table}

\subsection{Full Stock-Scale Metrics}
\label{app:full_scale_metrics}

Table~\ref{tab:main_scale_training_full_metrics} expands the compact
return-only scale table in the main text. The main table is intentionally terse,
so that the comparison across model sizes and SFT data views is easy to read.
Here we include the full set of backtest metrics for the same 2025
Ranked-Strategy stock benchmark: final portfolio value, Sharpe ratio, maximum
drawdown, trade count, win rate, and hold rate. These additional columns are
useful for distinguishing genuine return improvements from changes in trading
frequency or exposure. For example, two methods can have similar terminal returns
while reaching them through very different levels of turnover or interim loss.

\begin{table*}[t]
\centering
\scriptsize
\caption{Full MetaPS 2025 evaluation metrics under the Top10-Ranked input setting. Return, win rate, and hold rate are percentages; final value is shown in millions of dollars.}
\label{tab:main_scale_training_full_metrics}
\begin{tabular}{llrrrrrrr}
\toprule
Model & Variant & Return $\uparrow$ & Final Value $\uparrow$ & Sharpe $\uparrow$ & Max DD $\downarrow$ & Trades & Win Rate $\uparrow$ & Hold Rate \\
\midrule
MetaPS-0.8B & Base & 29.90 & 1.30M & 0.415 & 33.62 & 12 & 0.00 & 94.80 \\
 & V1 & 32.90 & 1.33M & 0.536 & 30.98 & 25 & 0.00 & 90.00 \\
 & V2 & 36.08 & 1.36M & 0.502 & 31.50 & 31 & 0.00 & 87.60 \\
 & V3 & \textbf{40.08} & 1.40M & 0.560 & 32.04 & 29 & 42.86 & 88.40 \\
 & Serial V1$\rightarrow$V2 & 39.54 & 1.40M & 0.486 & 32.32 & 34 & 33.33 & 86.40 \\
 & Serial V1$\rightarrow$V2$\rightarrow$V3 & 27.85 & 1.28M & 0.606 & 20.11 & 69 & 39.29 & 72.40 \\
\addlinespace
MetaPS-2B & Base & 28.50 & 1.29M & 0.494 & 32.29 & 14 & 100.00 & 94.40 \\
 & V1 & 28.28 & 1.28M & 0.666 & 30.78 & 29 & 28.57 & 88.40 \\
 & V2 & \textbf{38.98} & 1.39M & 0.459 & 32.33 & 34 & 33.33 & 86.40 \\
 & V3 & 36.09 & 1.36M & 0.701 & 24.85 & 61 & 23.81 & 75.60 \\
 & Serial V1$\rightarrow$V2 & 38.26 & 1.38M & 0.474 & 32.69 & 30 & 0.00 & 88.00 \\
 & Serial V1$\rightarrow$V2$\rightarrow$V3 & 34.73 & 1.35M & 0.601 & 8.60 & 94 & 30.77 & 62.40 \\
\addlinespace
MetaPS-4B & Base & 25.12 & 1.25M & 0.916 & 16.59 & 82 & 32.35 & 67.20 \\
 & V1 & 18.39 & 1.18M & 1.016 & 25.85 & 76 & 25.00 & 69.60 \\
 & V2 & 40.69 & 1.41M & 0.583 & 28.48 & 56 & 28.57 & 77.60 \\
 & V3 & 40.00 & 1.40M & 0.564 & 14.21 & 88 & 45.00 & 64.80 \\
 & Serial V1$\rightarrow$V2 & 33.26 & 1.33M & 0.581 & 29.47 & 54 & 30.77 & 78.40 \\
 & Serial V1$\rightarrow$V2$\rightarrow$V3 & \textbf{45.72} & 1.46M & 0.705 & 12.57 & 93 & 26.32 & 62.80 \\
\addlinespace
MetaPS-9B & Base & 32.69 & 1.33M & 0.762 & 27.30 & 43 & 20.00 & 82.80 \\
 & V1 & 45.39 & 1.45M & 0.721 & 14.15 & 85 & 37.84 & 66.00 \\
 & V2 & 31.24 & 1.31M & 0.690 & 27.53 & 51 & 25.00 & 79.60 \\
 & V3 & \textbf{50.29} & 1.50M & 0.668 & 15.29 & 89 & 31.43 & 64.40 \\
 & Serial V1$\rightarrow$V2 & 38.32 & 1.38M & 0.571 & 26.56 & 45 & 11.11 & 82.00 \\
 & Serial V1$\rightarrow$V2$\rightarrow$V3 & 40.68 & 1.41M & 0.522 & 26.64 & 48 & 23.53 & 80.80 \\
\bottomrule
\end{tabular}
\end{table*}

\subsection{Return-Curve Diagnostics}
\label{app:return_curves}

Aggregate return, Sharpe ratio, and maximum drawdown summarize a backtest into a
small number of scalars. Those scalars are necessary for comparison, but they can
hide how a method earns its return. A high terminal value may come from steady
compounding, from one short favorable interval, or from a late recovery after a
long underwater period. The following figures therefore plot the daily portfolio
state and related diagnostics over the held-out 2025 test year.

\begin{figure*}[!t]
    \centering
    \includegraphics[width=0.92\linewidth]{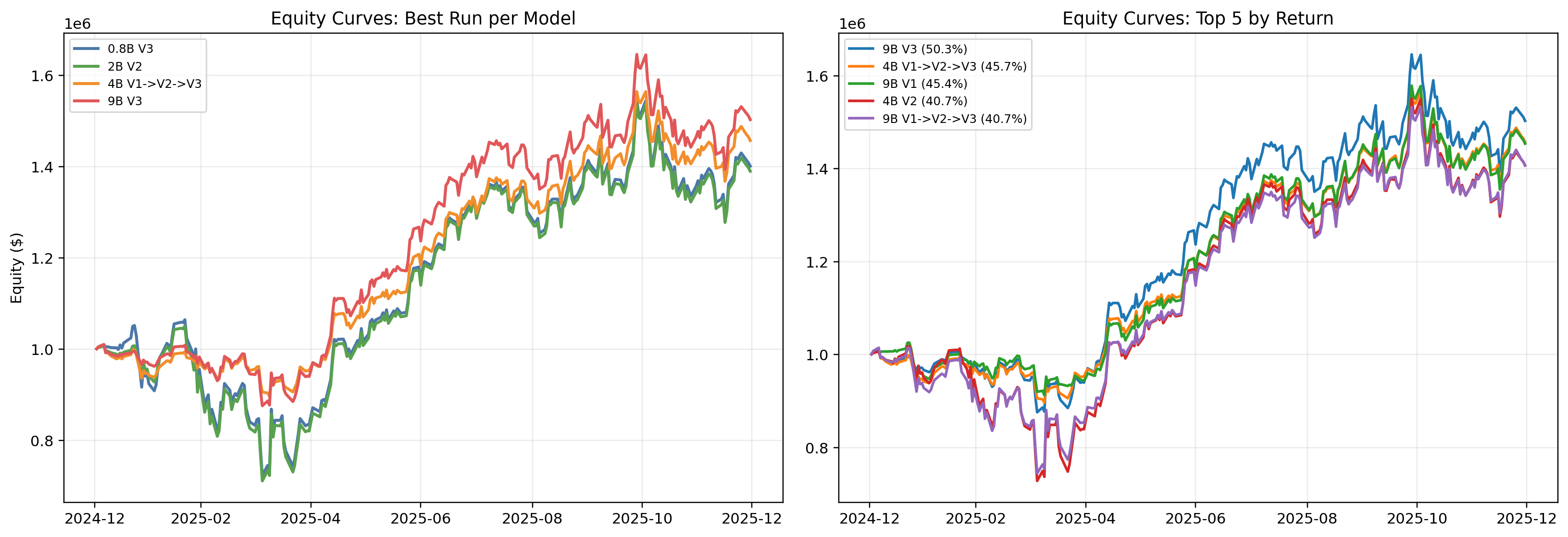}
    \caption{Portfolio equity curves for the main Qwen and MetaPS variants on
    the held-out 2025 stock benchmark. Each curve starts from the same initial
    capital and follows the daily backtester state, making it possible to see
    whether a method compounds steadily, recovers late, or depends on a small
    number of large moves.}
    \label{fig:appendix_equity_curves_2025}
\end{figure*}

Figure~\ref{fig:appendix_equity_curves_2025} first shows the equity curves for
the main Qwen and MetaPS variants. Because all curves are normalized to the same
initial capital, vertical separation directly reflects accumulated performance.
The shape of each curve is also informative: smooth upward movement indicates
persistent compounding, flat regions indicate long holding or inactive periods,
and abrupt changes reveal intervals in which the chosen strategy had especially
large influence on terminal return.

\begin{figure*}[!t]
    \centering
    \begin{subfigure}[t]{0.49\linewidth}
        \centering
        \includegraphics[width=\linewidth]{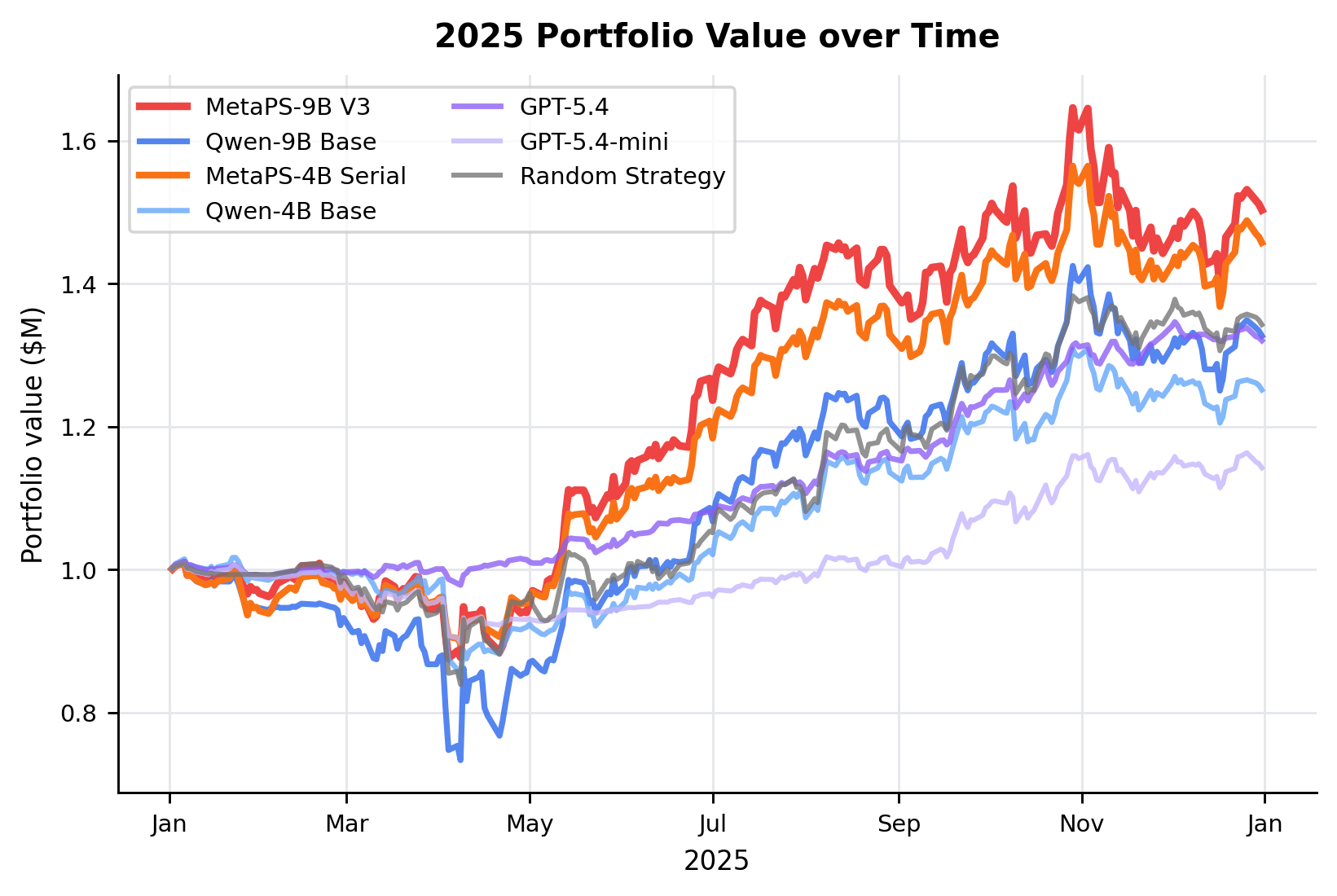}
        \caption{Portfolio-value trajectories.}
        \label{fig:return_equity_curves_baselines}
    \end{subfigure}
    \hfill
    \begin{subfigure}[t]{0.49\linewidth}
        \centering
        \includegraphics[width=\linewidth]{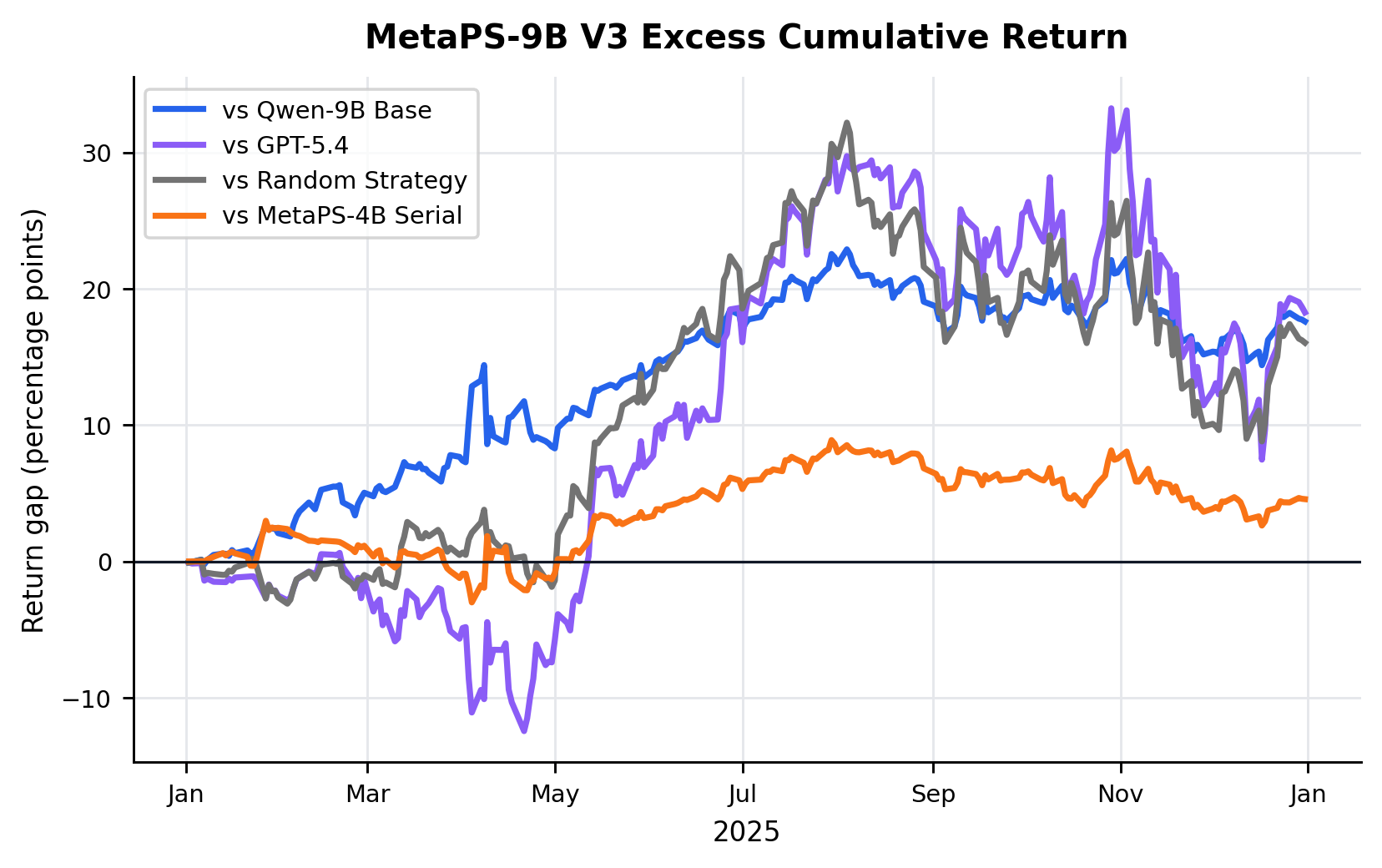}
        \caption{Cumulative return gap relative to MetaPS-9B V3.}
        \label{fig:return_excess_metaps9b}
    \end{subfigure}
    \caption{Return dynamics for MetaPS-9B V3 and representative baselines on
    the held-out 2025 stock benchmark. The equity-curve view shows the absolute
    portfolio paths, while the return-gap view emphasizes when MetaPS-9B V3
    gains or loses ground relative to each baseline. Positive gaps indicate that
    MetaPS-9B V3 has higher cumulative return up to that date.}
    \label{fig:combined_return_metrics}
\end{figure*}

Figure~\ref{fig:combined_return_metrics} widens the comparison to the
representative strategy-only, API, and Qwen baselines. The left panel reports the
same quantity as an equity curve, while the right panel re-expresses the result
as a cumulative return gap relative to MetaPS-9B V3. The gap plot is included
because absolute curves can be visually compressed when several methods move in
the same broad direction. Positive regions in the right panel indicate periods in
which MetaPS-9B V3 is ahead of a baseline; changes in slope show when that
advantage is being built or eroded.

\begin{figure*}[!t]
    \centering
    \begin{subfigure}[t]{0.49\linewidth}
        \centering
        \includegraphics[width=\linewidth]{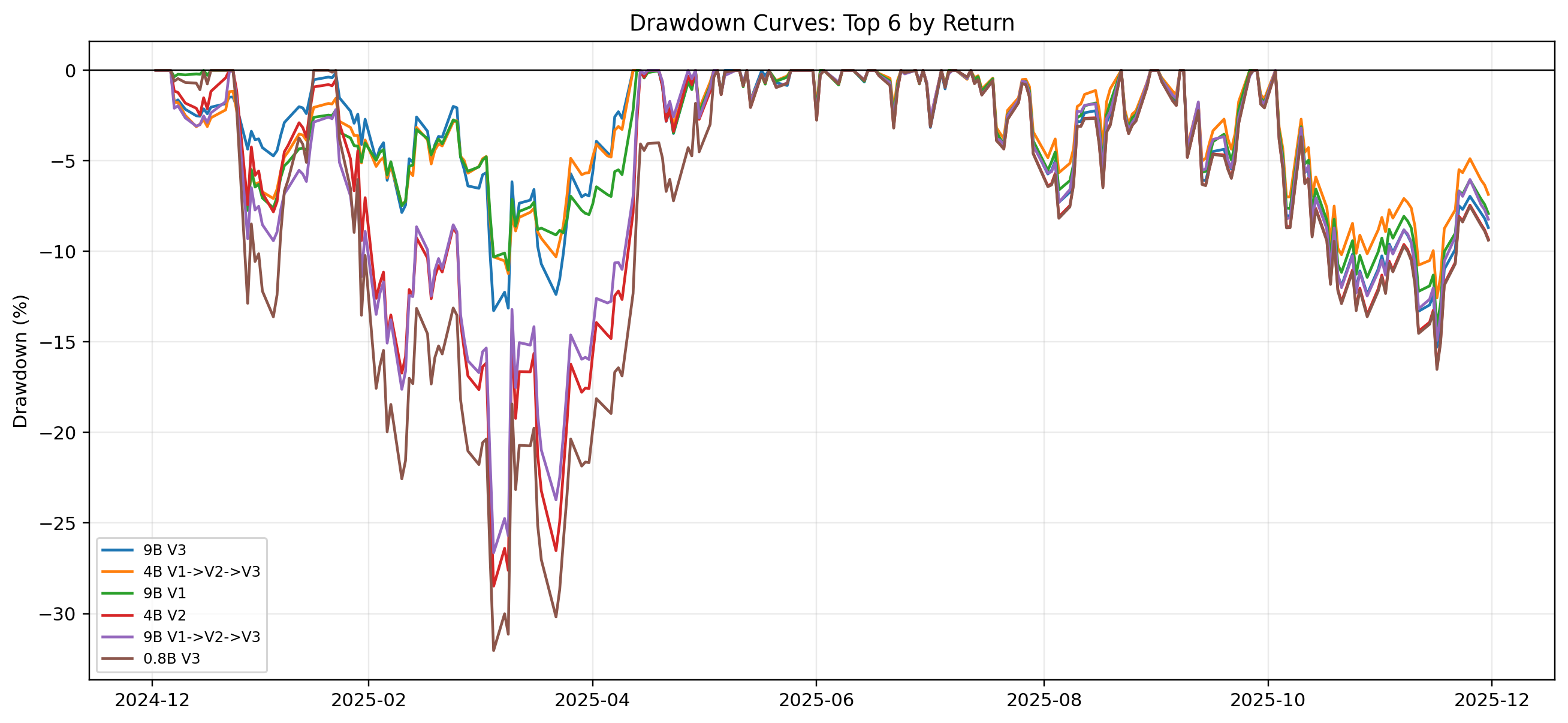}
        \caption{Drawdowns for the main stock benchmark variants.}
        \label{fig:appendix_drawdown_curves_2025}
    \end{subfigure}
    \hfill
    \begin{subfigure}[t]{0.49\linewidth}
        \centering
        \includegraphics[width=\linewidth]{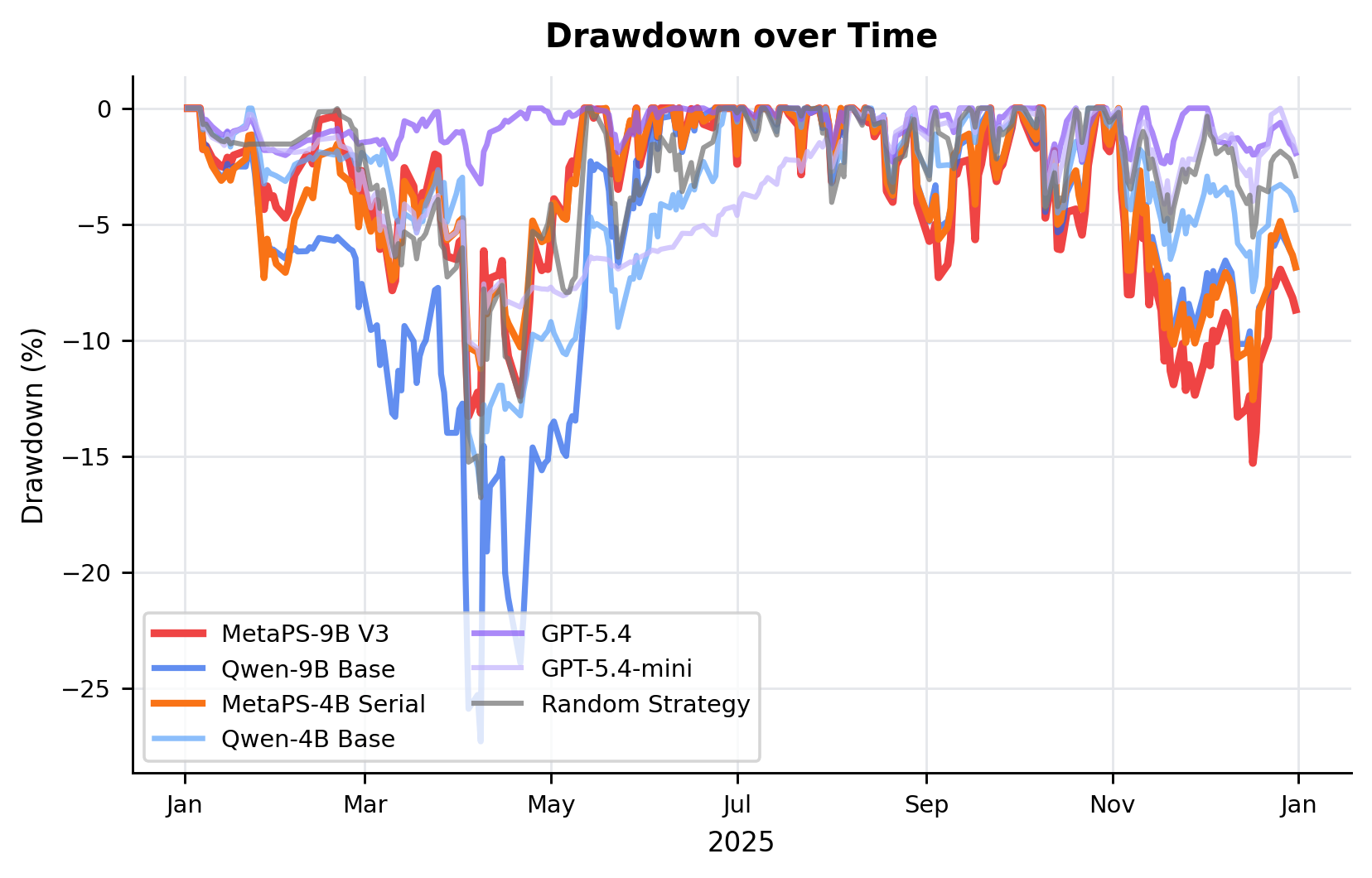}
        \caption{Drawdowns for representative baselines.}
        \label{fig:return_drawdown_baselines}
    \end{subfigure}
    \caption{Drawdown diagnostics on the held-out 2025 stock benchmark. Lower
    values indicate deeper declines from the running portfolio peak. Comparing
    the two panels helps separate methods that merely achieve high final return
    from those that also maintain a more stable path through the year.}
    \label{fig:combined_drawdown_metrics}
\end{figure*}

Return curves alone do not show the severity or duration of interim losses.
Figure~\ref{fig:combined_drawdown_metrics} therefore reports the corresponding
drawdown trajectories. Drawdown is measured relative to the running portfolio
peak, so deeper negative values indicate larger peak-to-trough losses and longer
periods below zero indicate slower recovery. This diagnostic complements both the
terminal return table and the Sharpe ratio by showing whether a method's gain is
achieved through a path that would require tolerating sustained losses.

\begin{figure*}[!t]
    \centering
    \begin{subfigure}[t]{0.49\linewidth}
        \centering
        \includegraphics[width=\linewidth]{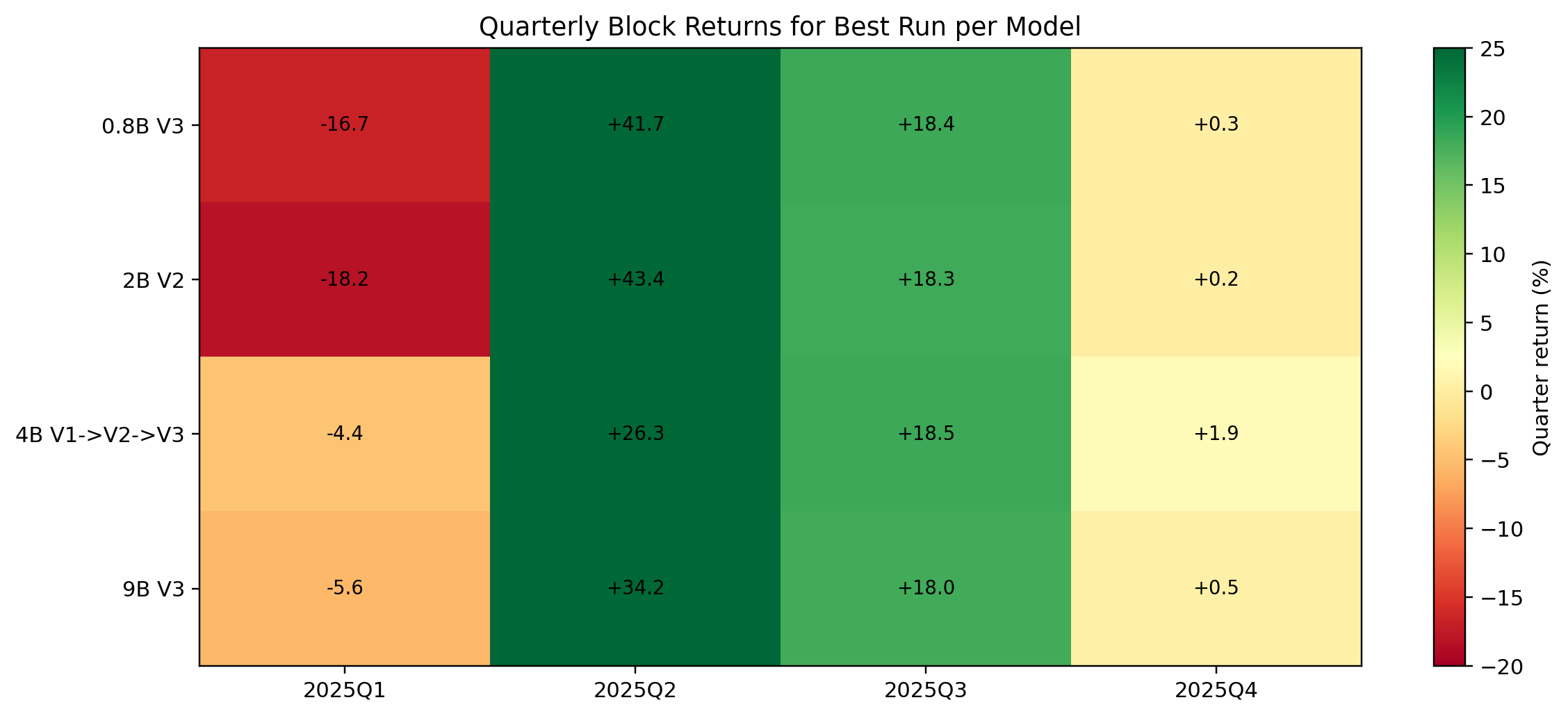}
        \caption{Block-level returns over the 2025 test year.}
        \label{fig:appendix_block_returns_2025}
    \end{subfigure}
    \hfill
    \begin{subfigure}[t]{0.49\linewidth}
        \centering
        \includegraphics[width=\linewidth]{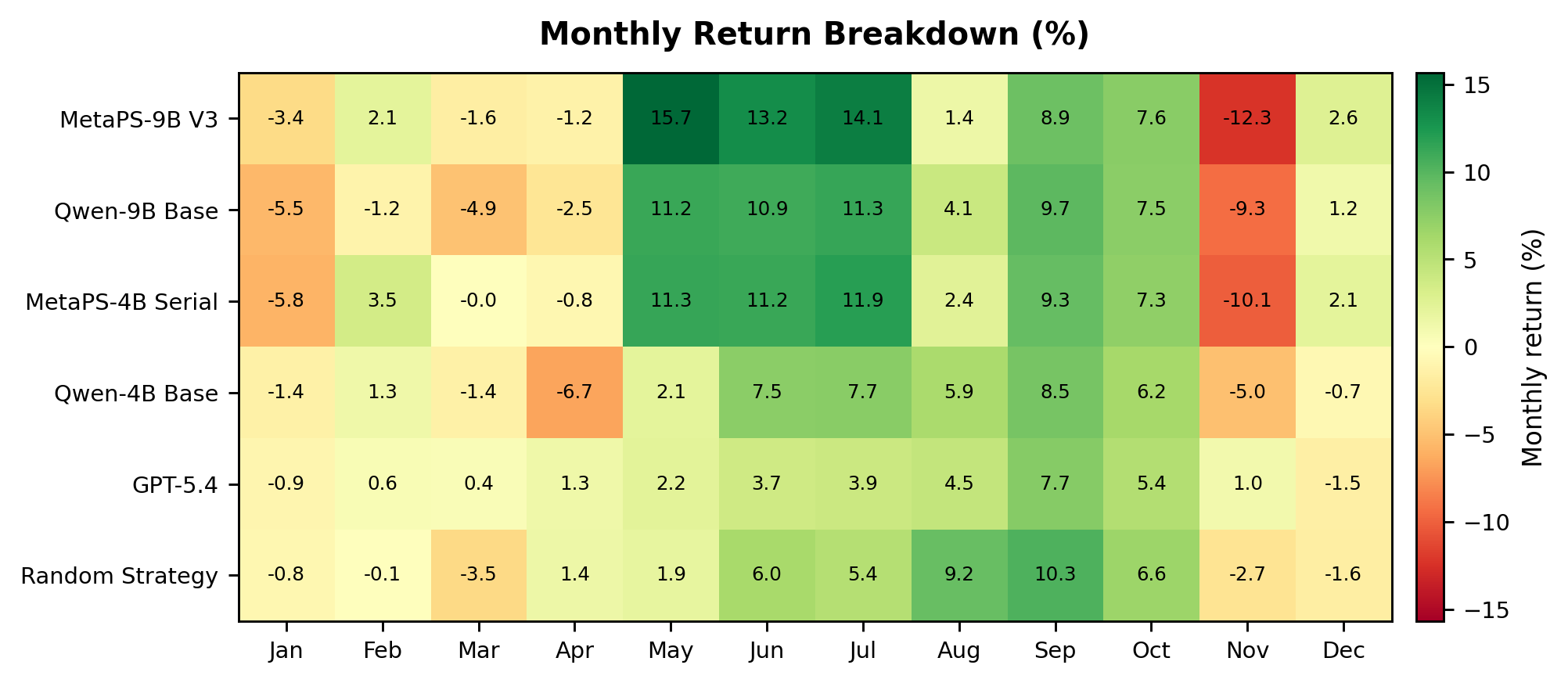}
        \caption{Monthly return breakdown.}
        \label{fig:return_monthly_heatmap}
    \end{subfigure}
    \caption{Calendar-level return diagnostics on the held-out 2025 stock
    benchmark. The block-level view summarizes performance over longer market
    segments, while the monthly heatmap shows whether gains and losses are
    concentrated in particular months.}
    \label{fig:combined_block_monthly}
\end{figure*}

Finally, Figure~\ref{fig:combined_block_monthly} aggregates the same test year
into coarser calendar views. The block-level plot asks whether the full-year
result is concentrated in one segment or persists across several contiguous
periods. The monthly heatmap provides a finer diagnostic of when each method is
adding or losing value. Together, these views make the temporal structure of the
benchmark more explicit and reduce the risk of over-interpreting a single
terminal-return number.

\FloatBarrier

\subsection{Additional Baseline and Ablation Figures}
\label{app:additional_baseline_figures}

\begin{figure*}[!t]
    \centering
    \begin{subfigure}[t]{0.48\linewidth}
        \centering
        \includegraphics[width=\linewidth]{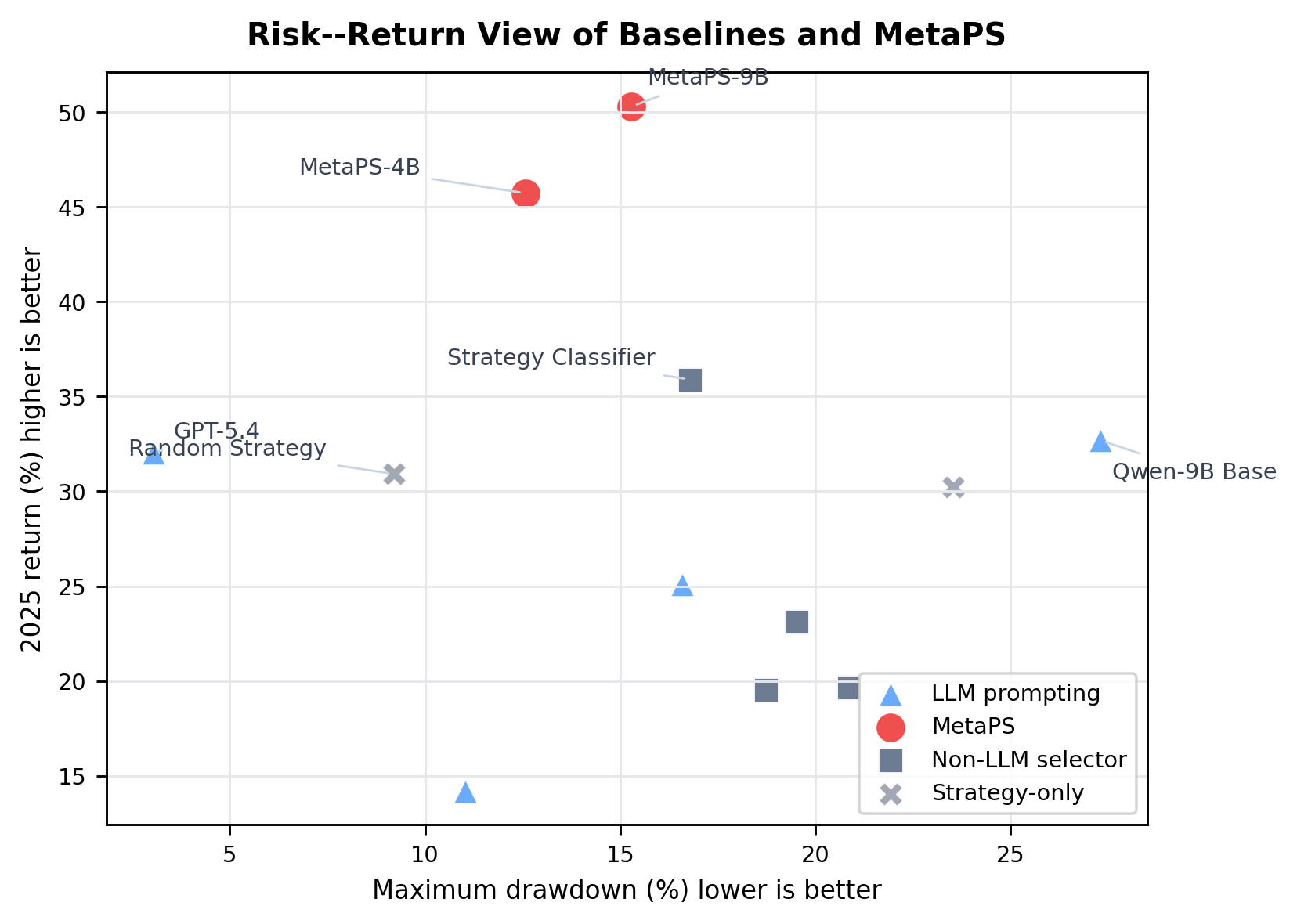}
        \caption{Risk--return comparison.}
        \label{fig:appendix_main_risk_return}
    \end{subfigure}
    \hfill
    \begin{subfigure}[t]{0.48\linewidth}
        \centering
        \includegraphics[width=\linewidth]{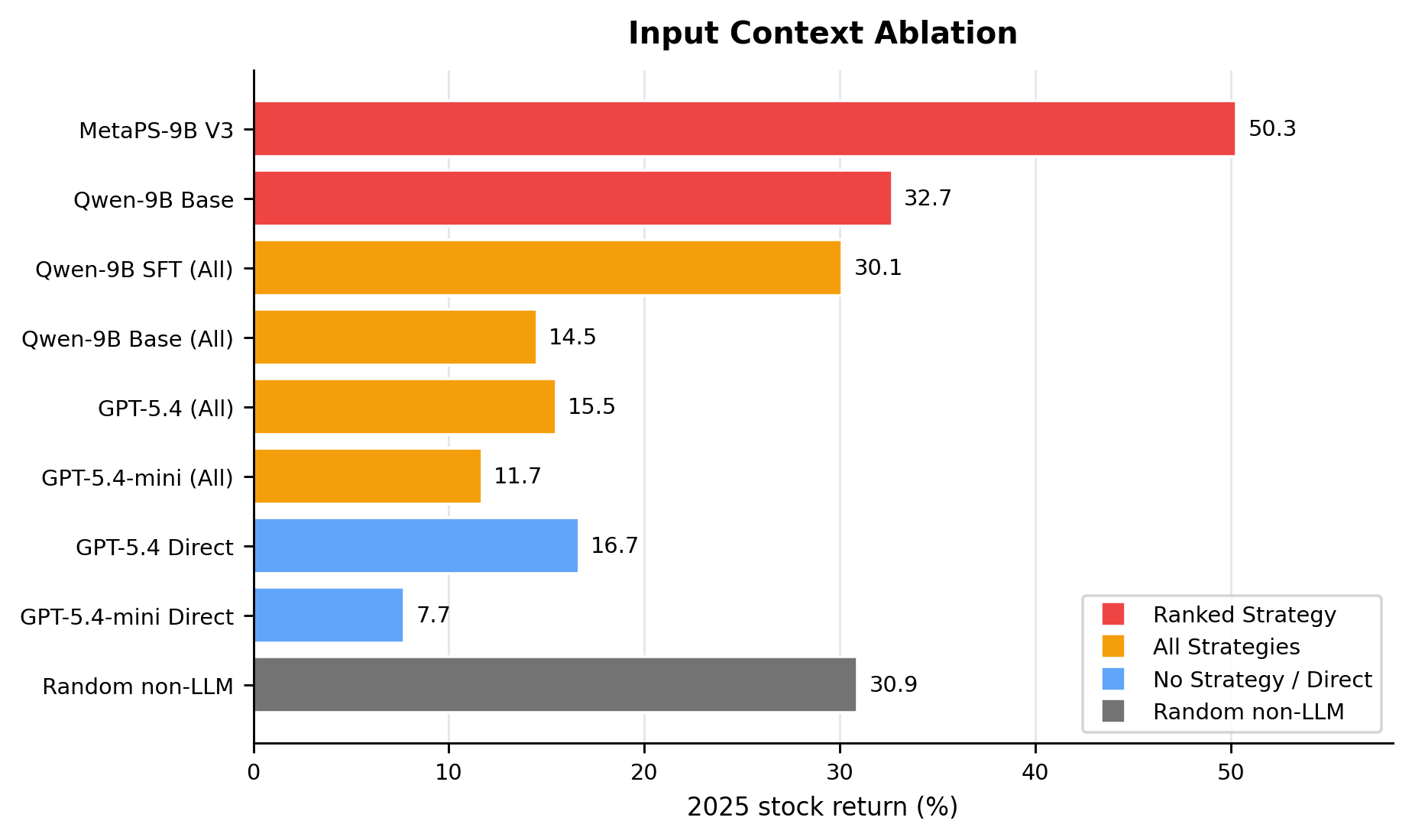}
        \caption{Input-context ablation.}
        \label{fig:appendix_context_ablation}
    \end{subfigure}
    \caption{Additional baseline diagnostics on the held-out 2025 stock
    benchmark. The left panel compares return against downside risk; the right
    panel summarizes how strategy-context information affects router behavior and
    performance.}
    \label{fig:appendix_baseline_pair}
\end{figure*}

The next set of figures connects the aggregate performance tables to more
interpretable experimental axes. Figure~\ref{fig:appendix_baseline_pair} places
methods in a risk--return plane and visualizes the input-context ablation from
Table~\ref{tab:llm_input_context}. The risk--return plot is useful because a
method that increases return only by accepting substantially larger drawdown
should be interpreted differently from one that improves both dimensions. The
context ablation, in turn, checks whether the router benefits from being given
strategy-level context rather than relying only on the raw market description.

\begin{figure*}[!t]
    \centering
    \begin{subfigure}[t]{0.49\linewidth}
        \centering
        \includegraphics[width=\linewidth]{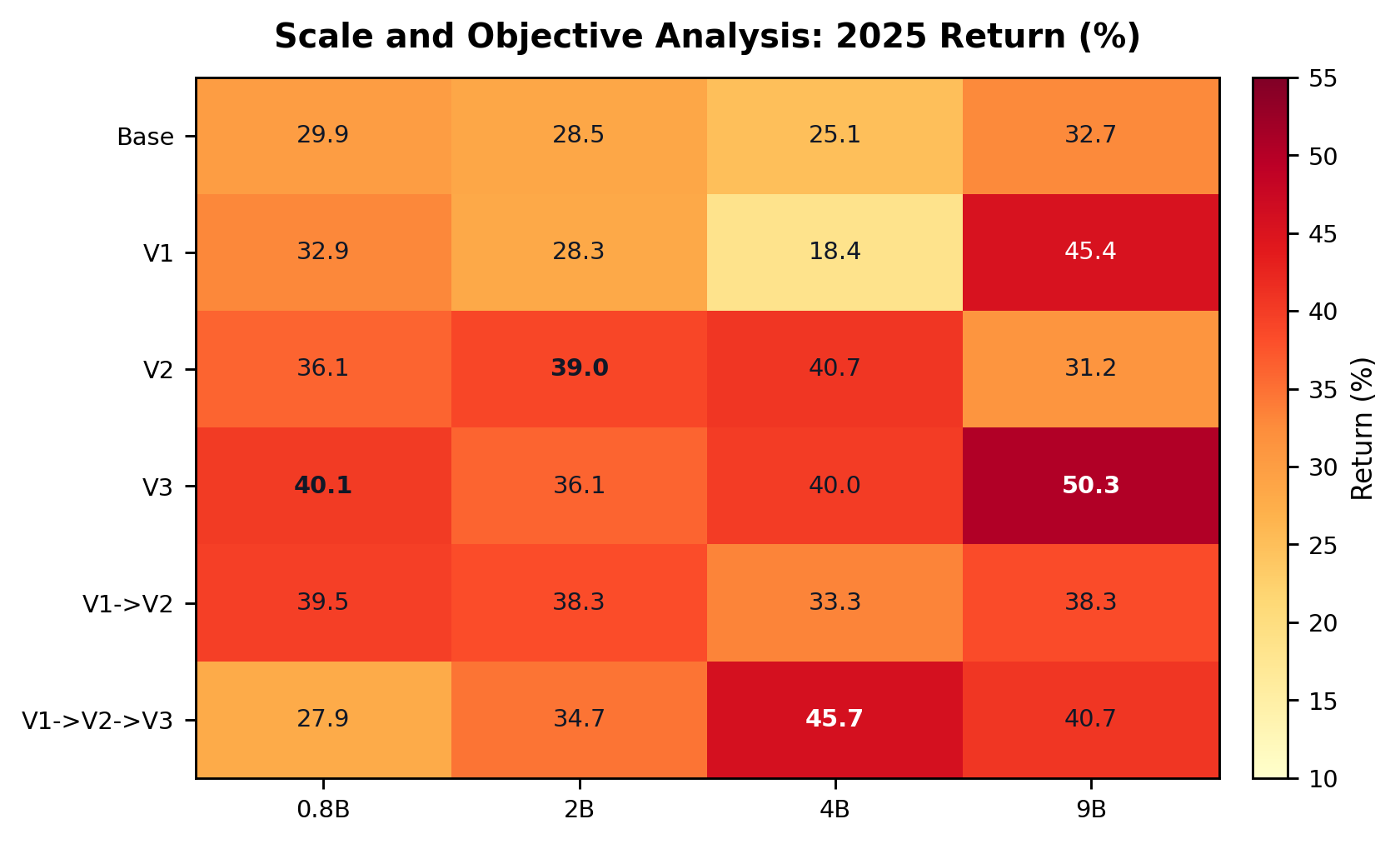}
        \caption{Return heatmap across model scales and training objectives.}
        \label{fig:appendix_scale_heatmap}
    \end{subfigure}
    \hfill
    \begin{subfigure}[t]{0.49\linewidth}
        \centering
        \includegraphics[width=\linewidth]{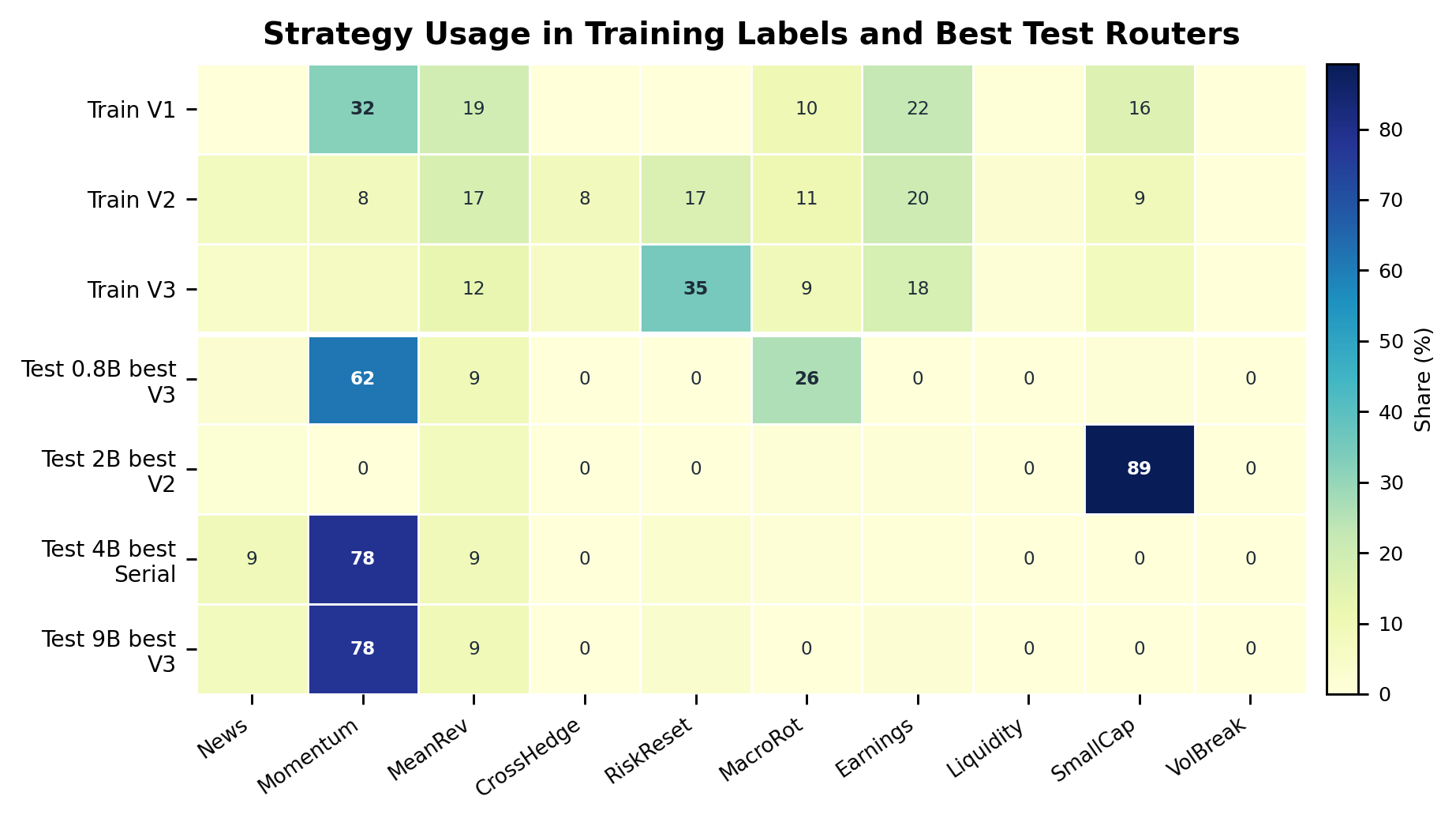}
        \caption{Strategy distributions for SFT labels and best test routers.}
        \label{fig:strategy_distribution_heatmap}
    \end{subfigure}
    \caption{Scale and behavior analysis for the 2025 stock benchmark. The left
    panel reports returns across Qwen model scales and MetaPS training
    objectives. The right panel reports the full ten-strategy distribution for
    SFT labels and the best test router at each model scale; values are row-wise
    percentages, and zero entries indicate strategies available to the router but
    not selected in that held-out run.}
    \label{fig:combined_scale_analysis}
\end{figure*}

Figure~\ref{fig:combined_scale_analysis} focuses on model scale and strategy
usage. The scale-by-objective heatmap summarizes how return changes across Qwen
model sizes and MetaPS data views, while the strategy-distribution heatmap shows
which programs are actually selected by the trained routers. This distinction is
important: an apparent scale effect in return is more informative when paired
with evidence that the larger or better-supervised router changes its strategy
allocation rather than simply reproducing the same behavior with a different
model size.

\FloatBarrier

\subsection{Behavior and Data-Distribution Diagnostics}
\label{app:behavior_diagnostics}

The preceding figures evaluate whether the routers perform well; this subsection
examines what they choose. Table~\ref{tab:strategy_behavior_best} reports the
cross-scale behavior summary behind Figure~\ref{fig:main_strategy_behavior}. For
each model size, it lists the best MetaPS variant, its BUY/HOLD/SELL action
counts, and its most frequently selected strategy programs. This table is meant
to make the learned policy more inspectable: two routers with similar returns
may still rely on different mixtures of strategies and therefore encode
different trading preferences.

\begin{table}[t]
\centering
\scriptsize
\caption{Strategy-use patterns of the best MetaPS variant at each model scale on the 2025 stock benchmark. Counts are over 250 daily decisions.}
\label{tab:strategy_behavior_best}
\setlength{\tabcolsep}{3pt}
\begin{tabular}{llrrrrp{0.35\linewidth}}
\toprule
Model & Variant & Ret. $\uparrow$ & Hold \% & Buy & Sell & Top selected strategies \\
\midrule
MetaPS-0.8B & V3 & 40.08 & 88.4 & 22 & 7 & momentum-follow (154), macro-rotation (65), mean-revert-fade (23) \\
MetaPS-2B & V2 & 38.98 & 86.4 & 31 & 3 & small-cap-breakout (223), mean-revert-fade (19), news-impulse (4) \\
MetaPS-4B & Serial V1$\rightarrow$V2$\rightarrow$V3 & 45.72 & 62.8 & 55 & 38 & momentum-follow (196), mean-revert-fade (22), news-impulse (22) \\
MetaPS-9B & V3 & 50.29 & 64.4 & 54 & 35 & momentum-follow (195), mean-revert-fade (23), news-impulse (19) \\
\bottomrule
\end{tabular}
\end{table}

The full strategy-distribution heatmap in
Figure~\ref{fig:strategy_distribution_heatmap} complements the compact main-text
behavior figure by showing all ten available strategies, including strategies
that are never selected by a particular held-out router. This is useful for
identifying whether a model is genuinely routing among multiple programs or has
collapsed onto a small subset of behaviors.

\begin{figure*}
    \centering
    \begin{subfigure}[t]{0.40\linewidth}
        \centering
        \includegraphics[width=\linewidth]{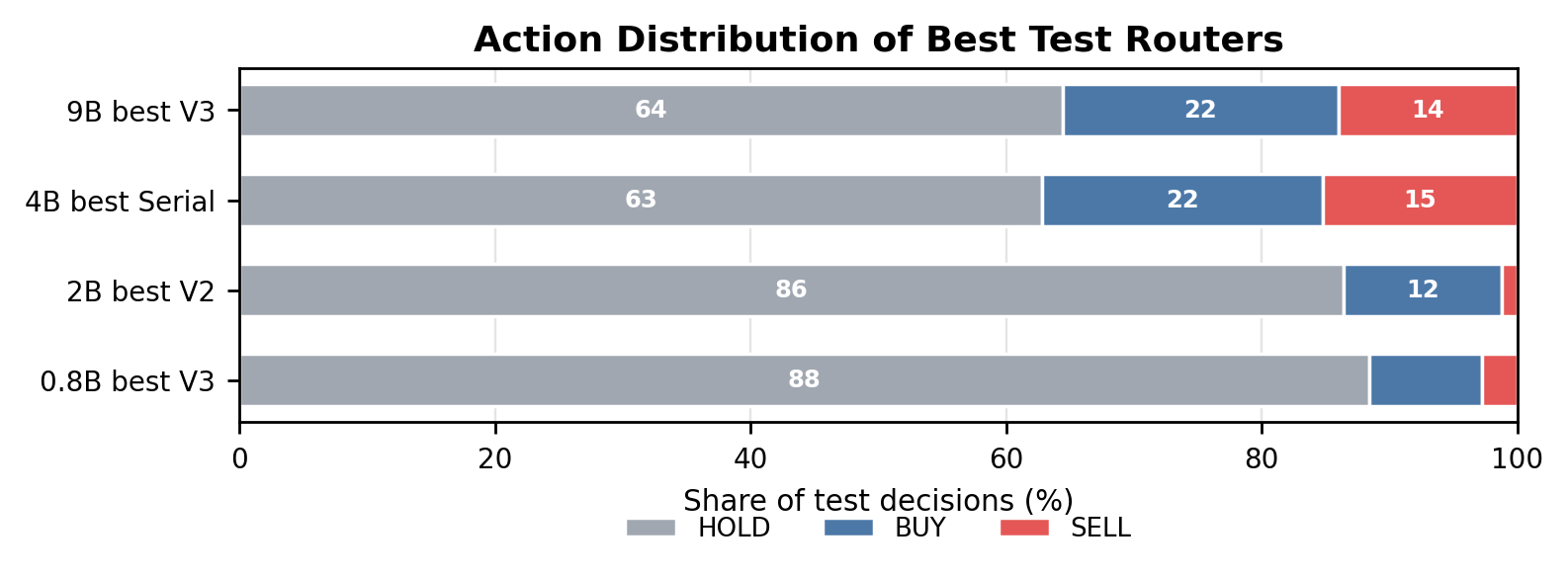}
        \caption{Best-router action distributions.}
        \label{fig:best_model_action_distribution}
    \end{subfigure}
    \hfill
    \begin{subfigure}[t]{0.60\linewidth}
        \centering
        \includegraphics[width=\linewidth]{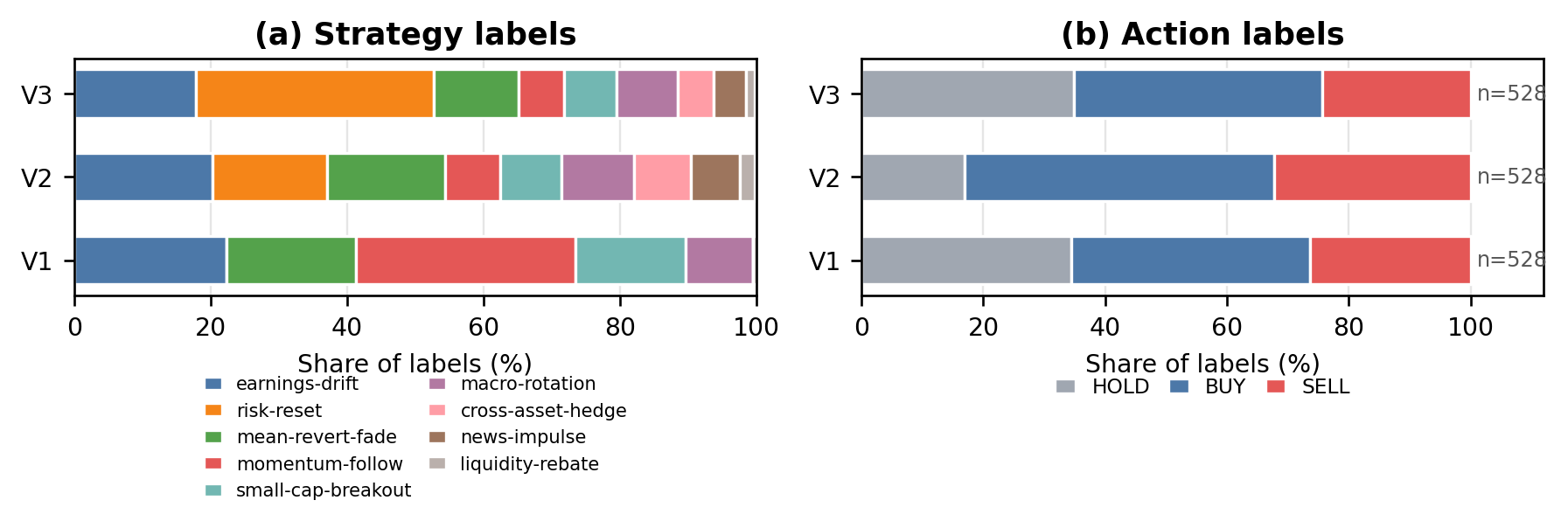}
        \caption{Training-label distributions.}
        \label{fig:training_label_distribution}
    \end{subfigure}
    \caption{Behavior diagnostics for realized action labels and SFT data views.
    The best-router distribution shows the downstream actions produced by the
    selected strategies, while the training-label distribution shows how the
    three SFT views differ before model training.}
    \label{fig:behavior_label_pair}
\end{figure*}

Figure~\ref{fig:behavior_label_pair} then moves one level downstream from
strategy choice to action labels. The left panel reports the BUY/HOLD/SELL
distribution induced by the best test router at each model scale. The right
panel reports how the three SFT data views differ before training. Each view
contains 528 examples, but the label distributions are intentionally different:
V1 is closer to short-horizon winner imitation, V2 shifts more mass toward active
BUY/SELL labels, and V3 rebalances the data to avoid relying on a single
aggressive behavior pattern. We keep this diagnostic in the appendix because
action type is downstream of the main strategy-selection question, but it helps
explain why different data views can produce different realized trading paths.

\begin{table}[t]
\centering
\small
\caption{Action-label distribution of the three MetaPS SFT data views.}
\label{tab:training_action_distribution}
\begin{tabular}{lrrrr}
\toprule
Data view & Samples & HOLD & BUY & SELL \\
\midrule
V1 & 528 & 182 (34.5\%) & 207 (39.2\%) & 139 (26.3\%) \\
V2 & 528 & 89 (16.9\%) & 269 (50.9\%) & 170 (32.2\%) \\
V3 & 528 & 184 (34.8\%) & 215 (40.7\%) & 129 (24.4\%) \\
\bottomrule
\end{tabular}
\end{table}

\FloatBarrier

\subsection{Sandbox Results}
\label{app:sandbox_results}

Table~\ref{tab:controlled_sandbox_app} reports the fuller controlled economic
sandbox comparison. The sandbox is not used as primary evidence for the stock
benchmark claims, because its dynamics are synthetic and do not reproduce the
full distributional complexity of real markets. Instead, it serves as a
controlled stress test of the interface: the same router chooses among shared
strategy programs, and the resulting terminal equity, regret, Sharpe ratio, and
drawdown are measured under a simpler environment where the source of gains and
mistakes is easier to inspect.

The pattern is consistent with the stock experiments. Learned selectors improve
over a random strategy baseline, API prompting baselines are competitive but do
not dominate the learned routers, and the best 4B MetaPS variants obtain the
strongest terminal equity and regret profile in this sandbox. Because the setting
is auxiliary, we report these results descriptively rather than treating them as
a replacement for the held-out 2025 stock benchmark.

\begin{table*}[!t]
\centering
\small
\caption{Auxiliary controlled economic sandbox results using shared
strategy-selection baselines, API prompting baselines, and 4B MetaPS routers.
The sandbox is reported as a controlled diagnostic rather than as primary stock
benchmark evidence.}
\label{tab:controlled_sandbox_app}
\resizebox{\textwidth}{!}{%
\begin{tabular}{lrrrrrr}
\toprule
Method & Terminal Equity $\uparrow$ & Return (\%) $\uparrow$
& Winner Share $\uparrow$ & Mean Regret $\downarrow$
& Sharpe $\uparrow$ & Max DD $\downarrow$ \\
\midrule
\rowcolor{gray!15}\multicolumn{7}{l}{\textbf{Strategy-only Baselines}} \\
Random Strategy & 13401 & 54.84 & 16.2 & 735.79 & 6.52 & 6.11 \\
Best Fixed Strategy & 18609 & 115.02 & 38.0 & 443.74 & 12.99 & 3.74 \\
\addlinespace[1pt]
\rowcolor{gray!15}\multicolumn{7}{l}{\textbf{Non-LLM Learned Selectors}} \\
Strategy Classifier & 26647 & 207.89 & 53.0 & 268.37 & 35.14 & 0.00 \\
Sequence Selector & 23872 & 175.84 & 50.0 & 339.97 & 18.52 & 1.21 \\
Reward-Ranking Selector & 26002 & 200.45 & 53.0 & 251.28 & 27.11 & 0.00 \\
RL Q-Policy Selector & 15030 & 73.67 & 41.0 & 680.20 & 8.22 & 5.17 \\
\addlinespace[1pt]
\rowcolor{gray!15}\multicolumn{7}{l}{\textbf{LLM and MetaPS Routers}} \\
Qwen-4B Base & 23879 & 175.92 & 41.0 & 399.45 & 18.25 & 1.21 \\
GPT-5.4-mini & 23249 & 168.64 & 54.0 & 236.77 & 25.08 & 0.01 \\
GPT-5.4 & 23554 & 172.16 & 53.0 & 251.29 & 17.42 & 1.52 \\
MetaPS-4B V1 & \textbf{31514} & \textbf{264.14} & \textbf{59.0} & \textbf{186.22} & \textbf{44.21} & \textbf{0.00} \\
MetaPS-4B V2 & 27631 & 219.26 & 57.0 & 230.85 & 18.69 & 1.07 \\
MetaPS-4B V3 & 30454 & 251.89 & \textbf{59.0} & 198.76 & 41.55 & 0.09 \\
\bottomrule
\end{tabular}
}
\end{table*}

\end{document}